\documentclass[nohyperref]{article}

\usepackage{microtype}
\usepackage{graphicx}
\usepackage{subfigure}
\usepackage{booktabs} 

\usepackage{hyperref}

\usepackage[accepted]{icml2022}

\usepackage{amsmath}
\usepackage{amssymb}
\usepackage{mathtools}
\usepackage{amsthm}
\usepackage{multirow}
\usepackage[capitalize,noabbrev]{cleveref}
\usepackage{colortbl}
\usepackage{stmaryrd}
\usepackage{pifont}
\usepackage{gensymb}
\usepackage{wrapfig,lipsum,booktabs}

\newcommand{\cmark}{\ding{51}}%
\newcommand{\xmark}{\ding{55}}%

\definecolor{dgreen}{rgb}{0.04,0.7,0.13}
\definecolor{maroon}{rgb}{0.75,0.07,0.03}
\definecolor{dblue}{rgb}{0.1,0.07,0.75}

\theoremstyle{plain}

\theoremstyle{definition}

\theoremstyle{remark}

\usepackage[textsize=tiny]{todonotes}

\usepackage{xspace}
\makeatletter
\DeclareRobustCommand\onedot{\futurelet\@let@token\@onedot}
\def\@onedot{\ifx\@let@token.\else.\null\fi\xspace}
\def\eg{\emph{e.g}\onedot} 
\def\ie{\emph{i.e}\onedot}

\def\wrt{w.r.t\onedot} 

\usepackage{makecell}

\usepackage{amssymb}
\usepackage{color}
\usepackage{xcolor}

\definecolor{lightgreen}{RGB}{100,220,100}
\definecolor{darkgreen}{RGB}{30,150,30}
\definecolor{darkblue}{RGB}{0,0,127}
\definecolor{darkyellow}{RGB}{171,133,0}
\definecolor{darkred}{RGB}{180,20,20}
\definecolor{darkmagenta}{RGB}{127,0,127}
\definecolor{darkcyan}{RGB}{0,127,127}

\definecolor{purple}{HTML}{9900ff}
\definecolor{darkpink}{HTML}{ff00ff}
\definecolor{maroon}{HTML}{980000}

\newcommand{\icmlcolor}[1]{\textcolor{mydarkblue}{#1}}

\usepackage{lipsum}
\usepackage{bbm}

\newcommand{\expectation}{\mathop{\mathbb{E}}}

\icmltitlerunning{Balancing Discriminability and Transferability for Source-Free Domain Adaptation}

\begin{document}

\twocolumn[
\icmltitle{Balancing Discriminability and Transferability \\ for Source-Free Domain Adaptation}



\icmlsetsymbol{equal}{*}

\begin{icmlauthorlist}
\icmlauthor{Jogendra Nath Kundu}{equal,iisc}
\icmlauthor{Akshay Kulkarni}{equal,iisc}
\icmlauthor{Suvaansh Bhambri}{equal,iisc}
\icmlauthor{Deepesh Mehta}{iisc}
\icmlauthor{Shreyas Kulkarni}{iisc}
\icmlauthor{Varun Jampani}{google}
\icmlauthor{R.\ Venkatesh Babu}{iisc}
\end{icmlauthorlist}

\icmlaffiliation{iisc}{Indian Institute of Science}
\icmlaffiliation{google}{Google Research}

\icmlcorrespondingauthor{J.\ N.\ Kundu}{jogendrak@iisc.ac.in}

\icmlkeywords{Machine Learning, ICML, Domain Adaptation, Computer Vision, Object Recognition, Semantic Segmentation}

\vskip 0.3in
]



\printAffiliationsAndNotice{\icmlEqualContribution} 

\begin{abstract}

Conventional domain adaptation (DA) techniques aim to improve domain transferability by learning domain-invariant representations; while concurrently preserving the task-discriminability knowledge gathered from the labeled source data. However, the requirement of simultaneous access to labeled source and unlabeled target renders them unsuitable for the challenging source-free DA setting. The trivial solution of realizing an effective original to generic domain mapping improves transferability but degrades task discriminability. Upon analyzing the hurdles from both theoretical and empirical standpoints, we derive novel insights to show that a mixup between original and corresponding translated generic samples enhances the discriminability-transferability trade-off while duly respecting the privacy-oriented source-free setting. A simple but effective realization\footnote[3]{\url{https://sites.google.com/view/mixup-sfda}} of the proposed insights on top of the existing source-free DA approaches yields state-of-the-art performance with faster convergence. Beyond single-source, we also outperform multi-source prior-arts across both classification and semantic segmentation benchmarks.

\end{abstract}

\section{Introduction}

\begin{figure}[ht]
\begin{center}
\centerline{\includegraphics[width=\columnwidth]{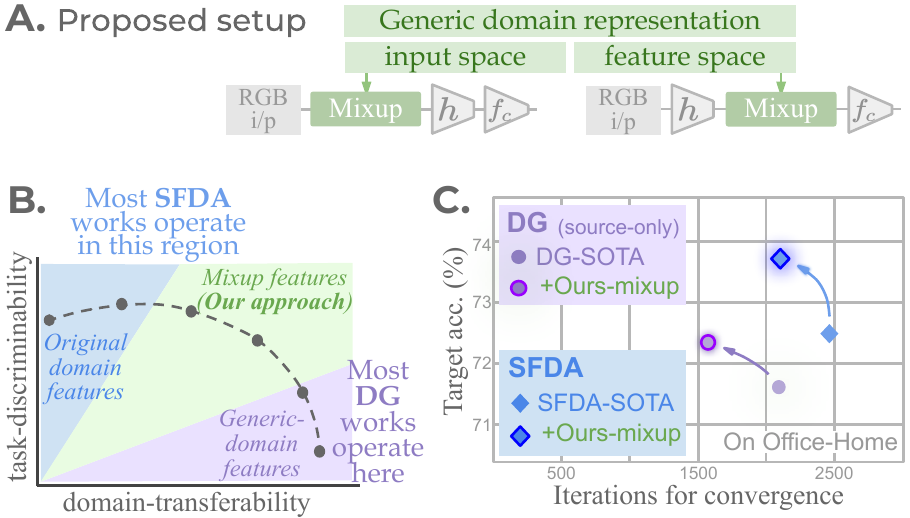}}
\vspace{-4mm}
\caption{\textbf{A.} We propose an instance-level mixup between original and generic-domain samples. \textbf{B.} Compared to previous SFDA (blue) and DG works (purple), we operate in an intermediate region (green) with an improved discriminability-transferability trade-off. \textbf{C.} As a result, our approach complements prior DG and SFDA SOTA works and produces better and faster convergence.
}
\label{fig:teaser}
\vspace{-12mm}
\end{center}
\end{figure}

Generally, in machine learning, it is assumed that the test samples are drawn from the same distribution as the training samples. However, in practice, a model often encounters a shift in the input distribution (\ie domain-shift), resulting in poor deployment performance. Unsupervised domain adaptation (DA) techniques seek to address this problem by transferring the task-discriminative knowledge from a labeled source domain to an unlabeled target domain. 

\vspace{-1mm}
From a theoretical viewpoint \cite{ben2006analysis}, the target error is upper bounded by three terms; \textbf{a)} the source error, \textbf{b)} the worst-case source-target domain discrepancy, and \textbf{c)} the error of a joint ideal hypothesis on source and target. With this line of thought, the widely adopted adversarial DA works \cite{ganin2016domain} aim to simultaneously minimize the first two terms using a feature extractor, task classifier, and domain classifier. The feature extractor is trained with two objectives, task classification on source and domain classifier fooling for source and target. While the first objective aims to preserve task-discriminability, the second objective aims to improve domain-transferability by encouraging domain-invariant representations. However, prior works \cite{chen2019transferability} show that the two objectives are at odds with each other \ie improving transferability leads to degraded discriminability and vice-versa. Certain works \cite{yang2020mind} explicitly address this by proposing ways to strike a suitable balance for better adaptation. But these works require joint access to both (labeled) source and (unlabeled) target data during training, making them unsuitable for source-free DA \cite{kundu2020towards}, where source and target data are not concurrently accessible. These DA scenarios with restricted data-sharing are privacy-oriented and thus, hold immense practical value.

\vspace{-1mm}
Recognizing this, we seek answers to following questions,
\textbf{1.} \textit{What are the key hurdles in extending the available techniques for the challenging source-free setting?} \\
\textbf{2.} \textit{What is the key design aspect to overcome these hurdles, thereby enabling effective source-free DA?}

In prior non-source-free works, the transferability is assessed as the inability of a domain classifier to segregate samples from the source and target domain. Clearly, such a realization is incompatible with the source-free constraint. Though task-discriminability can be gathered on the source-side (via supervised training on labeled source), it cannot be preserved while adapting to the unlabeled target in the absence of labeled source. In other words, the discriminability-transferability trade-off becomes a more severe problem that remains unaddressed even in available source-free techniques. As shown in Fig.\ \ref{fig:teaser}\icmlcolor{B}, source-free DA works aim to preserve the task-discriminability while domain generalization (DG) works \cite{li2017deeper} aim to achieve optimal transferability to generalize to unknown targets. Conversely, we believe an explicit effort to improve the trade-off (green area in Fig.\ \ref{fig:teaser}\icmlcolor{B}) could yield considerable improvements in source-free adaptation performance.

For the second question, we consider the idea of domain generic representations. A naive approach would be to develop \textit{source-to-generic} and \textit{target-to-generic} mappings which ideally leads to optimal transferability. In theory, the aim is to attain a generic representation that perfectly satisfies the following two criteria. First, the representation should be completely domain invariant. Second, the representation should not lose any task discriminative information. By definition, the discriminability-transferability trade-off would be well addressed. However, the key question remains, \textit{is such a representation realizable in practice?}

\begin{figure}[t]
\begin{center}
\centerline{\includegraphics[width=0.8\columnwidth]{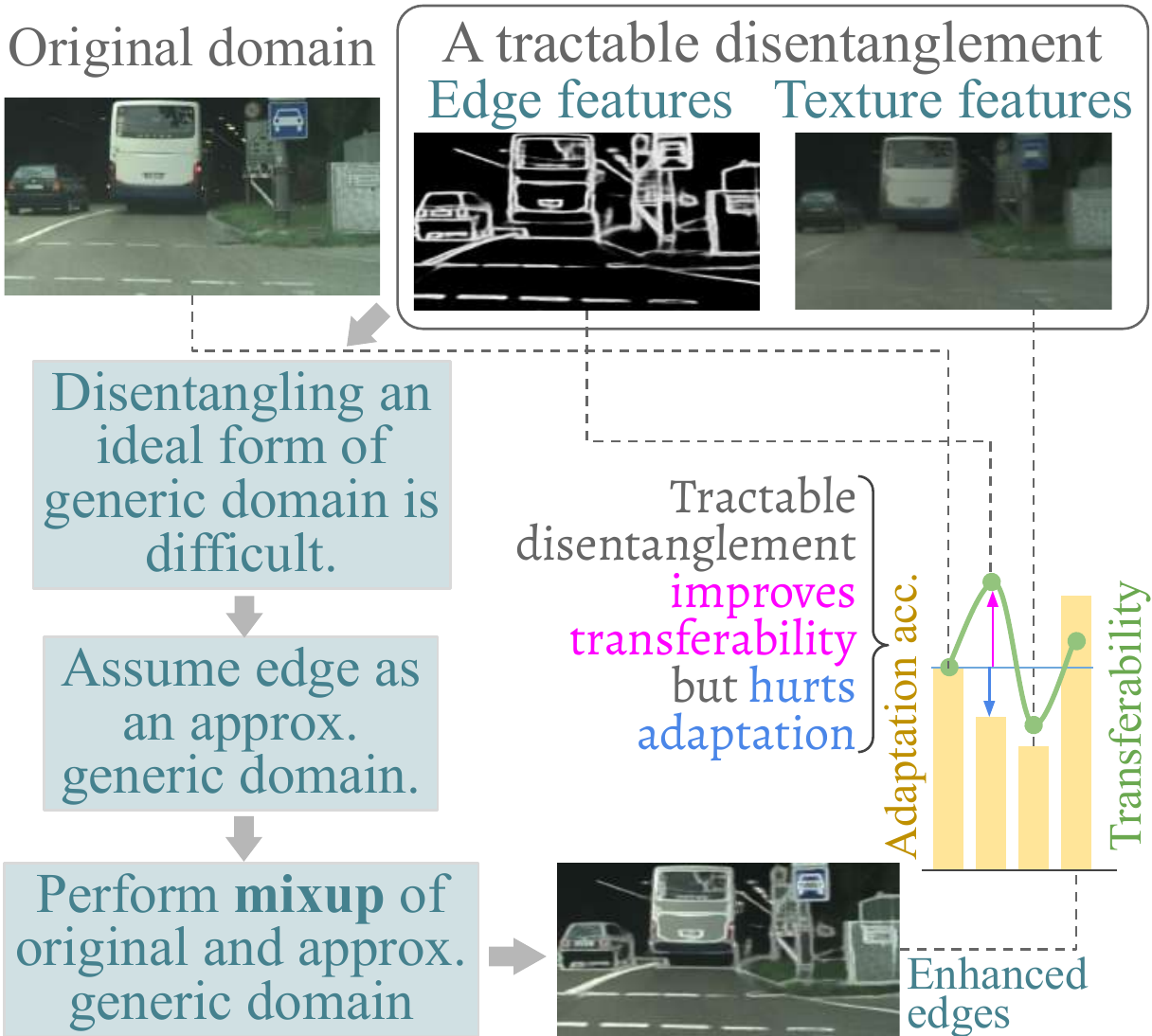}}
\vspace{-2mm}
\caption{
While an ideal generic domain is intractable, edges are a tractable disentanglement with higher transferability (pink arrow). The loss of task-discriminative information (blue arrow) is rectified by mixup between original and approximate generic domain.
}
\label{fig:theoretical_insight1}
\end{center}
\vspace{-8mm}
\end{figure}

To analyze this, we consider a synthetic to real-world road scene segmentation DA problem. The image samples for each domain can be considered an entangled mapping of two latent factors, \ie, \textbf{a)} Shape (edge-related features) and \textbf{b)} Texture (other than edge-related features). Such a disentanglement is easily achievable by segregating a simple edge representation from the original image through traditional contour estimation techniques (see Fig.\ \ref{fig:theoretical_insight1}). Now, we analyze whether the edge can be considered as a generic representation. We empirically observe (plot on the right of Fig.\ \ref{fig:theoretical_insight1}) that though edge satisfies the first criteria of domain invariance (higher transferability), it fails on the second criteria with degraded task discriminability (lower adaptation accuracy). Thus, it cannot be considered an ideal generic domain. This is because, while the texture varies across domains, it also holds crucial task discriminative features, \eg, differentiating between road, sidewalk, and hedges requires color/texture information. While we would ideally like to disentangle the purely task discriminative factors from the domain-related ones, the entanglement is highly non-linear. Implying it is challenging to realize ideal source-to-generic and target-to-generic mappings in practice.

Now, the key question that arises is, \textit{what could be a realizable approximation of the aforementioned ideal solution?} Following from the previous example, let us consider the edge representation as an approximate generic domain. Here, edge exhibits high domain-transferability with low task-discriminability (see Fig.\ \ref{fig:theoretical_insight1}). In comparison, the input representations of the original domains exhibit lower domain-transferability but higher task-discriminability. Aiming to attain an improved discriminability-transferability trade-off, we seek to develop an intermediate domain with a higher transferability than the original data domains and also a higher discriminability than the approximate generic domain. Following this overarching idea, we propose to realize the intermediate domain as an instance-level mixup of samples from the original and the approximate generic domains. This simple addition to existing source-free DA methods improves the implicit domain alignment, which in turn leads to better performance. The primary contributions of this work are:

\begin{itemize}

\vspace{-2mm}
\item 
We are the first to analyze the source-free DA problem from the perspective of discriminability-transferability trade-off. As a remedy to the low task discriminability of realizable generic domain, we operate on an intermediate mixup domain between the original and the realizable generic domain. We theoretically show that the mixup domains achieve a tighter bound on target error leading to improved adaptation performance.

\vspace{-2mm}
\item 
Based on our insights, we propose novel ways to realize approx.\ generic domains for mixup. We find that simple input-space edge representation better suits dense segmentation while feature-space mean-subdomain representation better suits non-dense classification.

\vspace{-2mm}
\item
This simple modification, on top of existing DA approaches, yields state-of-the-art performance across source-free benchmarks for single-source and multi-source DA on both classification and segmentation.

\end{itemize}

\section{Related Work}
\label{related_works}

\vspace{-1mm}
\textbf{Domain translation for DA.}
Adversarial alignment \cite{long2018conditional, rangwani2022closer} methods translate input domains to a domain-invariant representation to improve feature-space transferability. Input space adaptation methods \cite{murez2018image, russo2018source} explicitly translate \textit{source-to-target} or \textit{target-to-source} to improve transferability. Conversely, we propose separate source-to-generic and target-to-generic translations, realizable in both input and feature-space, to enable source-free DA.

\vspace{-1mm}
\textbf{Transferability-discriminability trade-off in DA.}
\citet{chen2019transferability} use spectral analysis and find that transferability resides in few top eigenvectors, while discriminability is spread across eigenvectors, which creates the trade-off. \citet{chen2020harmonizing} study this problem for object detection DA and propose hierarchical calibration of transferability. \citet{yang2020mind} propose autoencoder-based adversarial adaptation to avoid loss of discriminability.

\vspace{-1mm}
\textbf{Source-free DA.}
\citet{SHOT, SHOT++} use pseudo-labeling and information maximization to match target features with a fixed source classifier while \citet{morerio2020generative} generate target-style samples from a GAN. \citet{3C-GAN, NRC} focus on clustering-based regularization for source-free adaptation. \citet{kundu2020class} focuses on class-incremental SFDA. Apart from these, \citet{liu2021source}; \icmlcolor{Sivaprasad et al.} (\citeyear{sivaprasad2021uncertainty}); \citet{kundu2021generalize} particularly target segmentation-specific source-free DA.

\vspace{-1mm}
\textbf{Mixup.} Existing works interpolate different instances from the same class \cite{kim2021co}, different classes \cite{remix}, or even different domains \cite{FixBi} to better separate the class or domain clusters. They use mixup between distinct images where the label changes to a convex combination of labels and mixup images look unnatural. Conversely, our \textit{within-instance} mixup preserves the label (even for segmentation) with natural-looking mixup images.

\begin{figure}[t]
\begin{center}
\centerline{\includegraphics[width=0.9\columnwidth]{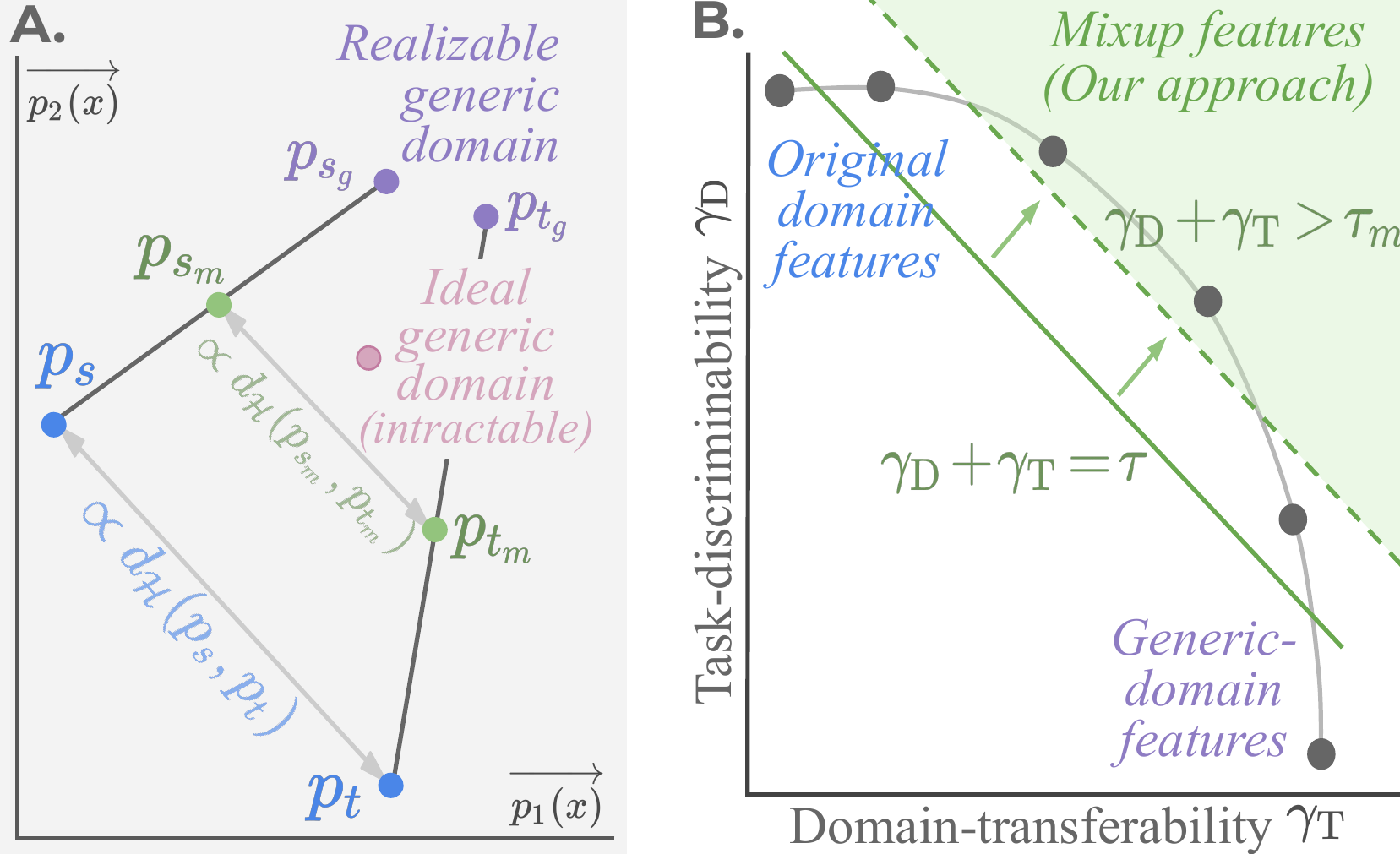}}
\vspace{-2mm}
\caption{\textbf{A.} $\overrightarrow{p_1(x)}$ and $\overrightarrow{p_2(x)}$ are some particular directions of variance in the affine space of marginal probability distributions \cite{krueger2021out}. Mixup of original domains with realizable generic domains yields mixup-domains with lower $d_\mathcal{H}$ (in green). \textbf{B.} Prior works lower bound $\gamma_\mathrm{D}+\gamma_\mathrm{T}$ by $\tau$ while our proposed mixup enables a better trade-off with a higher lower-bound $\tau_m$.
}
\label{fig:theoretical_insight2}
\end{center}
\vspace{-8mm}
\end{figure}

\vspace{-1mm}
\section{Approach}
\label{sec:approach}

\vspace{-1mm}
\noindent
\textbf{Problem setup.}
Under closed set DA, consider a labeled source dataset $\mathcal{D}_s = \{ (x_s, y_s): x_s\!\in\! \mathcal{X}, y_s \!\in\! \mathcal{C}\}$ where $\mathcal{X}$ is the input space and $\mathcal{C}$ denotes the label set. $x_s$ is drawn from the marginal distribution $p_s$. Also consider an unlabeled target dataset $\mathcal{D}_t = \{x_t: x_t \!\in\! \mathcal{X}\}$ where $x_t$ is drawn from the marginal distribution $p_t$. The task is to assign a label for each target image $x_t$ from the label set $\mathcal{C}$. Following \citet{ganin2016domain, long2015learning}, we use a backbone feature extractor $h\!:\!\mathcal{X}\!\to\! \mathcal{Z}$, where $\mathcal{Z}$ is an intermediate representation space, and a task classifier $f_c\!:\!\mathcal{Z}\!\to\! \mathcal{C}$. We operate under the source-free constraint \cite{3C-GAN} in a vendor-client paradigm \cite{kundu2020universal}. The vendor has the source dataset and can share a source-trained model with the client, without sharing the source data. The client adapts to the unlabeled target using the vendor-side model.

\vspace{-1mm}
\noindent
\textbf{Theoretical background.}
The expected source risk of the classifier $f_c$ with backbone $h$, with optimal labeling function $f_S\!:\!\mathcal{X}\!\to\!\mathcal{C}$, is $\epsilon_s(h) = \expectation_{x\sim p_s}[\mathbbm{1}(f_c\!\circ\! h(x) \neq f_S(x))]$, where $\mathbbm{1}(.)$ is the indicator function. Similarly, $\epsilon_t(h)$ is the expected target risk with optimal labeling function $f_T\!:\!\mathcal{X}\!\to\!\mathcal{C}$. \citet{ben2006analysis} present a theoretical upper bound on the expected target risk $\epsilon_t(h)$. For any backbone hypothesis $h\!\in\! \mathcal{H}$ with $\mathcal{H}$ as the hypothesis space and a domain classifier $f_d\!:\!\mathcal{Z}\!\to\!\{0, 1\}$ (0 for source, 1 for target),

{\footnotesize
\vspace{-4mm}
\begin{align}
    \begin{split}
    \epsilon_t(h) \leq &\;\epsilon_s(h) + d_\mathcal{H}(p_s, p_t) + \kappa(p_s, p_t), \text{ where}  \\
    d_\mathcal{H}(p_s,p_t)\!&=\!\sup_{h' \in \mathcal{H}} \left|\expectation_{p_s}[\mathbbm{1}(f_d \!\circ\! h'(x) \!=\! 1)] \!-\! \expectation_{p_t}[\mathbbm{1}(f_d \!\circ\! h'(x)\! =\! 1)]\right| \\
    \text{and } \kappa(p_s&, p_t) = \min_{h' \in \mathcal{H}} \epsilon_s(h') + \epsilon_t(h')
    \end{split}
    \label{eqn:target_risk_bound}
\end{align}
}
\vspace{-4mm}

\vspace{-2mm}
Here, $d_\mathcal{H}(p_s, p_t)$ denotes the $\mathcal{H}$-divergence that indicates the distribution shift between the source and target domains. $\kappa(p_s, p_t)$ represents the joint optimal error \ie the error of an ideal hypothesis on both source and target domains.

\vspace{-2mm}
\noindent
\textbf{Discriminability and transferability.}
\textit{Discriminability} refers to the {ease of separating different categories by a supervised classifier} trained on the features \cite{chen2019transferability}. \textit{Transferability} refers to the {invariance of feature representations across domains}. Exclusively improving the transferability leads to a drop in discriminability and vice versa \cite{chen2020harmonizing}. This is because the domain-specific information, that has to be removed to improve transferability, may contain entangled task-specific information important for discriminability. Based on the definitions, \citet{chen2019transferability} observe that the $\mathcal{H}$-divergence varies inversely with the transferability while the joint optimal error $\kappa(p_s, p_t)$ varies inversely with the discriminability of the backbone features. To quantify these two, we introduce a transferability metric $\gamma_\mathrm{T}$ and a discriminability metric $\gamma_\mathrm{D}$;

\vspace{-7mm}
\begin{align}
\begin{split}
    \gamma_\mathrm{T} = 1-d_\mathcal{H}(p_s, p_t); \;\;\;\;
    \gamma_\mathrm{D} = 1 - \frac{1}{2}\kappa(p_s, p_t)
\end{split}
\label{eqn:transf_disc_defn}
\end{align}
\vspace{-7mm}

For empirical computation of $\gamma_\mathrm{D}$ and $\gamma_\mathrm{T}$, we compute the expectation over the available domain samples. Note that $0\leq d_\mathcal{H}(., .)\leq 1$ while $0\leq \kappa(.,.)\leq 2$ (from Eq.\ \ref{eqn:target_risk_bound}).

We are the first to study the problem of handling transferability and discriminability in source-free DA. To analyze this problem, we provide an insight towards the question, \textit{what are the key hurdles in extending available discriminability-transferability based techniques for source-free DA?}

\noindent
\textbf{Insight 1. (Transferability-Discriminability in SFDA)}
\textit{The vendor model needs to achieve a good tradeoff between transferability and discriminability in order to be transferable to multiple clients, each posing a diverse target domain, while reasonably preserving the task discriminability.}

\noindent
\textbf{Remarks.}
In a vendor-client setting, the client desires a model with good discriminability in the target domain \ie good task performance on target data. However, the client cannot significantly influence the discriminability of a vendor-side model without labeled data. Thus, the vendor needs to preserve the discriminability for the clients while improving the transferability to serve multiple future clients \ie a good tradeoff benefits both vendor and client.

\vspace{-1mm}
\noindent
\textbf{Domain translation without concurrent access.}
While existing domain translation methods are limited by the concurrent access constraint, a domain translation method can be devised to be employed separately on source and target without data sharing. As per Insight \icmlcolor{1}, vendor needs to improve transferability to serve multiple clients. So, a naive solution would be to translate source or target to a \textit{generic-domain} with marginal distribution $p_{s_g}$ or $p_{t_g}$, where features possess both high transferability and high discriminability. However, in practice, these \textit{source-to-generic} and \textit{target-to-generic} translations would result in a loss of discriminability as domain-specific and task-specific information is entangled in the original domains (see Fig.\ \ref{fig:theoretical_insight1}). Thus, to balance domain-specificity and task-specificity, we propose mixup \cite{mixup} between original domain samples and corresponding generic-domain translated samples. Formally, \textit{source-to-mixup} translation in the input-space is,

\vspace{-5mm}
\begin{equation}
    x_{s_m} = \lambda x_{s_g} + (1-\lambda) x_s
    \label{eqn:mixup_input}
\end{equation}
\vspace{-4mm}

\vspace{-2mm}
\noindent
where $x_{s_g}$ are the generic-domain samples (effectively drawn from the hypothetical $p_{s_g}$) corresponding to the original domain samples $x_s$. And $x_{s_m}$ is the source-mixup sample (effectively drawn from the hypothetical $p_{s_m}$), while the mixup ratio $\lambda$ is a scalar constant. The equivalent equation for \textit{target-to-mixup} translation is $x_{t_m} = \lambda x_{t_g} + (1-\lambda) x_t$.

\begin{figure*}[t]
\vspace{-2mm}
\begin{center}
\centerline{\includegraphics[width=\textwidth]{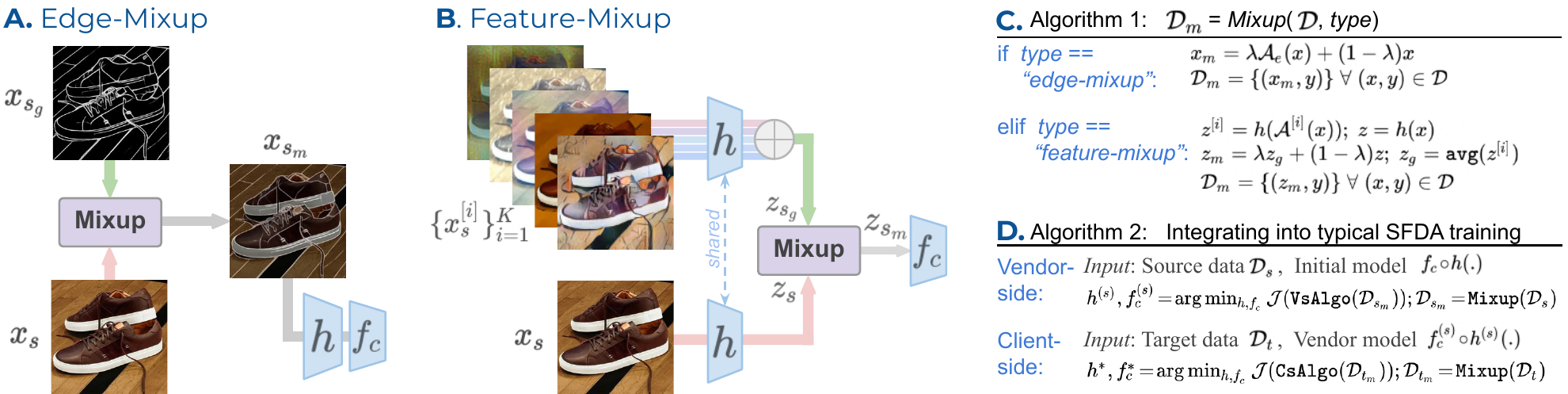}}
\vspace{-3mm}
\caption{\textbf{A.} Edge-mixup involves mixup of an input sample $x_s$ and the corresponding edge representation $x_{s_g}$ (Sec.\ \icmlcolor{3.2.1}). \textbf{B.} Feature-mixup involves mixup of the input features $z_s=h(x_s)$ and corresponding generic domain features $z_{s_g}$, obtained as the feature mean of augmented sub-domain samples $\{x_s^{[i]}\}_{i=1}^K$ (Sec.\ \icmlcolor{3.2.2}). \textbf{C.} Algorithm for edge-mixup and feature-mixup to convert an original dataset $\mathcal{D}$ into a mixup dataset $\mathcal{D}_m$. \textbf{D.} Algorithm to integrate our proposed mixup into a typical wrapper for source-free DA training.}
\label{fig:approach}
\end{center}
\vspace{-8mm}
\end{figure*}

\vspace{-2mm}
\subsection{Theoretical insights}

Now, we theoretically analyze the impact of using mixup distributions instead of original distributions w.r.t. Eq. \ref{eqn:target_risk_bound}.

\vspace{-1mm}
\textbf{Insight 2. (Effect of mixup on $\boldsymbol{d_\mathcal{H}}$ and $\boldsymbol{\kappa}$)}
\textit{We postulate that, keeping the same hypothesis space $\mathcal{H}$, the overall discriminability and transferability for the mixup distributions will be upper bounded by that for the original distributions,}

\vspace{-6mm}
\begin{equation}
    d_\mathcal{H}(p_{s_m}, p_{t_m}) + \kappa(p_{s_m}, p_{t_m}) \leq d_\mathcal{H}(p_{s}, p_{t}) + \kappa(p_{s}, p_{t})
    \label{eqn:transf_disc_bound}
\end{equation}
\vspace{-6mm}

\textbf{Remarks.}
Most prior works \cite{saito2018maximum} attempt to minimize these two terms by optimizing on the $\mathcal{H}$-space. Conversely, we highlight the perspective of manipulating the input distributions. Prior works lower bound $\gamma_\mathrm{D}+\gamma_\mathrm{T}$ by some threshold $\tau$ (solid green line in Fig.\ \ref{fig:theoretical_insight2}\icmlcolor{B}). Whereas by using the mixup distributions, the lower bound can be increased to $\tau_m$ (dotted green line in Fig.\ \ref{fig:theoretical_insight2}\icmlcolor{B}). This leads to a better tradeoff between transferability and discriminability. Consequently, mixup achieves a tighter upper bound (Eq.\ \ref{eqn:target_risk_bound}) for $\epsilon_t(h)$ since the remaining term $\epsilon_s(h)$ can be minimized easily in both cases with the supervised source data.

\vspace{-1mm}
\noindent
\textbf{a) Transferability for mixup vs. original distributions.}
To support Insight \icmlcolor{2}, we first analyze the relation between the $\mathcal{H}$-divergences for the mixup and original distributions. Intuitively, the samples of mixup distributions $p_{s_m}$ or $p_{t_m}$ contain less domain-specific information compared to the original domain samples from $p_s$ or $p_t$. Thus, the cross-domain feature transferability of $p_{s_m}$ or $p_{t_m}$ should improve. Now, we present a theoretical result to support our intuition.

\vspace{-1mm}
\noindent
\textbf{Theorem 1.\ (Mixup $\boldsymbol{\mathcal{H}}$-divergence)}
\textit{Assume that original source $p_s$ and target $p_t$ are easily separable} \ie \textit{perfect accuracy for domain classifier $f_d$. Also assume that source-generic $p_{s_g}$ and target-generic $p_{t_g}$ domains are impossible to separate} \ie \textit{accuracy of $f_d$ imitates that of a random classifier. For a linear domain classifier $f_d$;}

\vspace{-6mm}
\begin{equation}
    d_\mathcal{H}(p_{s_m}, p_{t_m})\leq d_\mathcal{H}(p_{s}, p_{t})
    \label{eqn:hdiv_bound}
\end{equation}
\vspace{-6mm}

\vspace{-1mm}
\noindent
\textbf{Remarks.}
We make three assumptions in this proof. First, original domains should be easily separable, giving perfect accuracy for the domain classifier $f_d$ used in $d_\mathcal{H}$ (see Eq. \ref{eqn:target_risk_bound}). This is reasonable since realistic domain shifts are found to be easily separable for empirical domain classifiers (even for linear classifiers \ie our second assumption). Finally, we assume that the source-generic and target-generic domains are impossible to separate. This is also a fair assumption since a generic-domain has high transferability by definition. Proof is provided in {Appendix \ref{app:subsec:theorem1_proof}}. We illustrate the same result of lower mixup $\mathcal{H}$-divergence in Fig.\ \ref{fig:theoretical_insight2}\icmlcolor{A}.

\vspace{-1mm}
\noindent
\textbf{b) Discriminability for mixup vs. original distributions.}
As discussed earlier, the discriminability varies inversely with the joint optimal error $\kappa(p_s, p_t)$. Since transferability and discriminability are at odds with each other, the lower $d_\mathcal{H}(p_{s_m}, p_{t_m})$ should increase $\kappa(p_{s_m}, p_{t_m})$ \wrt $\kappa(p_s, p_t)$. However, mixup with corresponding generic-domain samples implies that the task discriminative features will be preserved to a good extent since the common characteristics across the original and generic-domain samples are preserved. Unlike Theorem \icmlcolor{1}, it is not possible to theoretically arrive at a result relating $\kappa(p_{s_m}, p_{t_m})$ and $\kappa(p_{s}, p_{t})$, as the assumptions on $f_c$ become unrealistic (see {Appendix \ref{app:subsec:mixupk_relation}}). However, as long as task discriminative features are preserved (this is observed for low $\lambda$ values empirically in Fig.\ \ref{fig:analysis}\icmlcolor{A}), mixup discriminability $\kappa(p_{s_m}, p_{t_m})$ cannot drop w.r.t. the original $\kappa(p_{s}, p_{t})$ and Eq.\ \ref{eqn:transf_disc_bound} holds. Thus, mixup provides an improved transferability-discriminability tradeoff.

\vspace{-1mm}
\noindent \textbf{3.1.1. Criteria to realize generic-domain mixup}

\vspace{-1mm}
There are two aspects involved in realizing the generic-domain mixup. First, \textit{how to obtain generic-domain samples corresponding to original samples?} The ideal generic-domain $p_{s_g}$ or $p_{t_g}$ has features with high domain-transferability. Thus, the criteria for choosing a original-to-generic domain mapping would be the effectiveness of the translation technique in disentangling the domain-generic characteristics. This can be measured through the transferability metric $\gamma_\mathrm{T}$ for source and target generic-domain samples. We empirically observe (Fig.\ \ref{fig:analysis}\icmlcolor{A}) that our proposed generic domains ($\lambda\!=\!1$) exhibit higher $\gamma_\mathrm{T}$.

Second, \textit{how to perform the mixup?} In the proposed method, we perform mixup of generic-domain samples with corresponding original samples as a convex combination with a fixed mixup ratio $\lambda$ (see Eq.\ \ref{eqn:mixup_input}). While we show the performance to be insensitive over a wide range of $\lambda$ values (see Fig.\ \ref{fig:analysis}\icmlcolor{C}), there may be other ways to perform mixup. For instance, we may employ a learnable combination via meta-learning \cite{wei2021metaalign}. However, we choose the simple convex combination, which validates our theoretical insights. Other options can be explored in future work.

\begin{table*}[t]
    \centering
    \vspace{-4mm}
    \setlength{\tabcolsep}{12pt}
    \caption{\textbf{Single-Source Domain Adaptation (SSDA)} on Office-31 and VisDA benchmarks. SF indicates \textit{source-free} adaptation and \textcolor{darkgreen}{\textbf{(+x.x)}} indicates absolute improvements over the corresponding baseline methods NRC and SHOT++ (current source-free SOTA).}
    \label{tab:ssda_office31_visda}
    \resizebox{\textwidth}{!}{
    \begin{tabular}{lccccccccll}
        \toprule
        \multirow{2}{20pt}{\centering Method}& \multirow{2}{*}{\centering SF} 
        &&  \multicolumn{7}{c}{\textbf{Office-31}
        } & \multicolumn{1}{c}{\textbf{VisDA}} \\
        \cmidrule(lr){4-10}\cmidrule(lr){11-11}
        &&& A$\rightarrow$D & A$\rightarrow$W & D$\rightarrow$W & W$\rightarrow$D & D$\rightarrow$A & W$\rightarrow$A & Avg & S $\rightarrow$ R \\
        \midrule
		CDAN+RADA~\cite{RADA} & \xmark && 96.1 & 96.2 & 99.3 & 100.0 & 77.5 & 77.4 & 91.1 & 76.3\\ 
		FAA~\cite{FAA} & \xmark && 94.4 & 92.3 & 99.2 & 99.7 & 80.5 & 78.7 & 90.8 & 82.7\\ 
		FixBi~\cite{FixBi} & \xmark& & 95.0 & 96.1 & 99.3 & 100.0 & 78.7 & 79.4 & 91.4 & 87.2 \\
		\midrule
		HCL~\cite{HCL} & \cmark && 90.8 & 91.3 & 98.2 & 100.0 & 72.7 & 72.7 & 87.6 & 83.5\\ 
		CPGA~\cite{CPGA} & \cmark && 94.4 & 94.1 & 98.4 & 99.8 & 76.0 & 76.6 & 89.9 & 84.1\\ 
		${\text{A}^{2}\text{Net}}$~\cite{A2Net} & \cmark && 94.5 & 94.0 & 99.2 & 100.0 & 76.7 & 76.1 & 90.1 & 84.3\\ 
		VDM-DA~\cite{VDM-DA}  & \cmark && 93.2 & \textbf{94.1} & 98.0 & 100.0 & 75.8 & 77.1 & 89.7 & 85.1\\ 
		NRC~\cite{NRC} & \cmark && 96.0 & 90.8 & 99.0 & 100.0 & 75.3 & 75.0 & 89.4 & 85.9\\ 
		\rowcolor{gray!10} \textit{Ours (edge-mixup)} + NRC & \cmark && 96.1 & 92.4 & 99.2 & 100.0 & 76.9 & 77.1 & 90.3 \textcolor{lightgreen}{(+0.9)} & 86.4  \textcolor{lightgreen}{(+0.5)} \\
		
		\rowcolor{gray!10} \textit{Ours (feat-mixup)} + NRC & \cmark && \textbf{96.3} & 92.8 & \textbf{99.2} & 100.0 & 77.4 & 77.5 & 90.5 \textcolor{darkgreen}{\textbf{(+1.1)}} & {87.3} \textcolor{darkgreen}{\textbf{(+1.4)}} \\
		SHOT++~\cite{SHOT++} & \cmark && 94.3 & 90.4 & 98.7 & 99.9 & 76.2 & 75.8 & 89.2 & {87.3}\\ 

		\rowcolor{gray!10} \textit{Ours (edge-mixup)} + SHOT++ & \cmark && 94.4 & 92.0 & 98.9 & 100.0 & 77.8 & 77.9 & 90.2 \textcolor{lightgreen}{(+1.0)} & 	87.5 \textcolor{lightgreen}{(+0.2)} \\
		
		\rowcolor{gray!10} \textit{Ours (feat-mixup)} + SHOT++ & \cmark && 94.6 & 93.2 & 98.9 & \textbf{100.0} & \textbf{78.3} & \textbf{78.9} & \textbf{90.7} \textcolor{darkgreen}{\textbf{(+1.5)}} & 	\textbf{87.8} \textcolor{darkgreen}{\textbf{(+0.5)}} \\
        \bottomrule
    \end{tabular}
    }
\vspace{-4mm}
\end{table*}

\vspace{-2mm}
\subsection{Training algorithm and realizable generic-domains}

\vspace{-1mm}
We aim to demonstrate that our proposed mixup is complementary to existing DA methods (both source-free and non-source-free). To this end, we simply replace the original training datasets $\mathcal{D}_s$ and $\mathcal{D}_t$ in existing methods with our devised mixup datasets $\mathcal{D}_{s_m}$ and $\mathcal{D}_{t_m}$ respectively. Note the use of only the devised datasets, different from \textit{data augmentation} which would also use the original datasets.

\vspace{-1mm}
\textbf{Vendor-side training} (Fig.\ \ref{fig:approach}\icmlcolor{C}).
Consider an existing vendor-side training algorithm $\texttt{VsAlgo}(\mathcal{D}_s)$ that trains the backbone $h$ and classifier $f_c$ on the original source dataset $\mathcal{D}_s$. We minimize the objective of the same training algorithm but on the source-mixup dataset $\mathcal{D}_{s_m}$. Formally,

\vspace{-4mm}
\begin{equation}
    \min_{h, f_c} \mathcal{J}(\texttt{VsAlgo}(\mathcal{D}_{s_m}))
\end{equation}
\vspace{-4mm}

where $\mathcal{J}(.)$ represents the objective or loss function.

\vspace{-1mm}
\textbf{Client-side training} (Fig.\ \ref{fig:approach}\icmlcolor{D}).
Consider an existing client-side training algorithm $\texttt{CsAlgo}(\mathcal{D}_t)$. Similar to the vendor-side training, we train $h$ and $f_c$ on the target-mixup dataset $\mathcal{D}_{t_m}$ using the objective of $\texttt{CsAlgo}$. Formally,

\vspace{-4mm}
\begin{equation}
    \min_{h, f_c} \mathcal{J}(\texttt{CsAlgo}(\mathcal{D}_{t_m}))
\end{equation}
\vspace{-4mm}

Next, we discuss an input-space generic-domain representation to obtain the mixup datasets $\mathcal{D}_{s_m}$ and $\mathcal{D}_{t_m}$. Following  this, we devise a more flexible feature-space generic-domain representation that emulates the expected properties.

\noindent
\textbf{3.2.1. Edge representation as a generic-domain}

\noindent
\textbf{Motivation.}
General image classification or semantic segmentation tasks are based primarily on recognizing object shapes. Thus, any candidate representation for a generic-domain needs to possess at least the shape information. Intuitively, edge representation satisfies this criteria as it preserves the shape information while removing domain-variant information like color or texture.

\vspace{-1mm}
\textbf{Datasets.}
Consider the labeled source-edge-mixup dataset $\mathcal{D}_{s_m}\!=\!\{(x_{s_m}, y_s)\!:\!x_{s_m} \!=\! \lambda x_{s_g} + (1-\lambda) x_s, (x_s, y_s) \in \mathcal{D}_s \}$ where $\mathcal{D}_s$ is the original source dataset and $x_{s_g}=\mathcal{A}_e(x_s)$ is the edge representation of $x_s$. We use \citet{soria2020dexined} for the edge detector $\mathcal{A}_e(.)$.
Similarly, the unlabeled target-edge-mixup dataset is $\mathcal{D}_{t_m}=\{ x_{t_m} \}$.

\noindent
\textbf{3.2.2. Feature-space generic-domain representation}

\textbf{Motivation.}
While edge representation is intuitive as a generic-domain, it poses some limitations. First, edges may not represent a generic-domain for all possible tasks. For instance, in texture classification, shape is not the primary discriminative characteristic. Here, edge-mixup would hinder the discriminability more than it assists the transferability. Second, while an input-space generic-domain (like edges) seems intuitive for some tasks, it may be very challenging to devise the same for other tasks. To address these limitations, we introduce a more flexible feature-space generic-domain.

\vspace{-1mm}
\textbf{Augmented sub-domains.}
We perform the mixup in the feature-space of the backbone $h$. Consider the features $z_s = h(x_s)$ for a source sample $x_s$. We choose a set of task-preserving image augmentations $\{ \mathcal{A}^{[i]}(.) \}_{i=1}^K$ from a pool of commonly used augmentations. The augmentation should not distort task discriminative factors while significantly modifying domain variant factors. For example, a strong stylization technique like AdaIN (\icmlcolor{Huang et al.,} \citeyear{huang2017adain}) satisfies this criteria. Whereas weaker augmentations like Gaussian blurring are not chosen, as they fail to significantly alter the domain variant factors. Intuitively, manipulating the domain-variant factors simulates a sample from an \textit{augmented sub-domain}. See {Appendix \ref{app:subsec:experimental_settings}} for more details.

\vspace{-1mm}
Concretely, we extract a set of $K$ features $\{z_s^{[i]}\!:\! z_s^{[i]}\! =\! h(x_s^{[i]})\}_{i=1}^K$ where $x_s^{[i]} = \mathcal{A}^{[i]}(x_s)$ is the $i^\text{th}$ augmentation of $x_s$.
Since each of the augmented sub-domain features represents the same task discriminative features, but vary in domain-variant features, we use the feature mean as the generic-domain features, \ie $z_{s_g}=\frac{1}{K}\sum_{i=1}^K z_s^{[i]}$.

Usually, DA works \cite{saito2018maximum, rangwani2021submodular} apply domain alignment or other losses at feature-level as deep features exhibit high domain discrepancy \cite{stephenson2021on, kundu2022amplitude}. Thus, different augmentations of the same instance exhibit high feature-level domain-variance. Then, feature mean (generic-domain) effectively diffuses this variance while preserving task-related factors.

\vspace{-1mm}
\textbf{Datasets.} Consider labeled source-feature-mixup dataset $\mathcal{D}_{s_m} \!=\! \{ (z_{s_m}, y_s): z_{s_m} \!=\! \lambda z_{s_g} + (1-\lambda)z_s, (x_s, y_s) \in \mathcal{D}_s \}$ where $z_s\! =\! h(x_s)$. Similarly, unlabeled target-feature-mixup dataset is $\mathcal{D}_{t_m} \!=\! \{ z_{t_m} \}$. Note that using the same augmentations for both source and target does not require concurrent source-target access \ie does not violate the source-free constraints. We re-use the dataset notations from edge-mixup to avoid introducing more notations.

\vspace{-1mm}
During training, the backbone $h$ is only updated through the computation graph of $z_s$, \ie the gradients do not flow through the computation graph of $z_{s_g}$. This is because the gradients corresponding to each augmentation may be at odds with each other and should not be aggregated directly.
Also note that the final adapted model is finetuned on the original target data using the client-side algorithm \ie $\texttt{CsAlgo}(\mathcal{D}_t)$, for a small number of iterations, to remove the mixup requirement at inference-time. This ensures fair comparisons with prior arts using the same architecture.

\section{Experiments}
\label{sec:experiments}

\vspace{-1mm}
We thoroughly assess our technique against numerous state-of-the-art methods in different DA scenarios.

\begin{table*}[t]
    \vspace{-4mm}
    \centering
    \caption{\textbf{Multi-Source Domain Adaptation (MSDA)} on DomainNet and Office-Home. * indicates results from released code. We outperform \textit{source-free} (SF) prior arts despite not using domain labels. \textcolor{darkgreen}{\textbf{(+x.x)}} indicates improvements over the source-free SOTA NRC.}
    \label{tab:msda_compare}
    \setlength{\tabcolsep}{3pt}
    \resizebox{1\linewidth}{!}{%
        \begin{tabular}{lcccccccclcccccl}
            \toprule
            \multirow{2}{30pt}{\centering Method} & \multirow{2}{*}{\centering SF} & \multirow{2}{*}{\parbox{3cm}{\centering \vspace{4pt} w/o Domain \\ Labels}} & \multicolumn{7}{c}{\textbf{DomainNet}
            } && \multicolumn{5}{c}{\textbf{Office-Home}} \\
            \cmidrule(lr){4-10} \cmidrule(lr){12-16}
            & & &$\shortrightarrow$C & $\shortrightarrow$I & $\shortrightarrow$P & $\shortrightarrow$Q & $\shortrightarrow$R & $\shortrightarrow$S & Avg & & $\shortrightarrow$Ar & $\shortrightarrow$Cl & $\shortrightarrow$Pr & $\shortrightarrow$Rw & Avg \\
            \midrule
            {SImpAl$_{50}$~\cite{SImpAl}} & \xmark & \xmark &   66.4 & 26.5 & 56.6 & 18.9 & 68.0 & 55.5 & 48.6 & & 70.8 & 56.3 & 80.2 & 81.5 & 72.2 \\ 
            {CMSDA~\cite{CMSDA}} & \xmark & \xmark &  70.9 & 26.5 & 57.5 & 21.3 & 68.1 & 59.4 & 50.4 & &  71.5 & 67.7 & 84.1 & 82.9 & 76.6 \\ 
            {DRT \cite{DRT}} & \xmark & \xmark &   71.0 & 31.6 & 61.0 & 12.3 & 71.4 & 60.7 & 51.3 &  & - & - & - & - & -\\ 
            STEM~\cite{STEM} & \xmark & \xmark & 72.0 & 28.2 & 61.5 & 25.7 & 72.6 & 60.2 & 53.4 & & - & - & - & - & -  \\ 
            \midrule
            {Source-combine} &\xmark & \cmark &57.0 &23.4 &54.1 &14.6 &67.2 &50.3 & 44.4 & & 58.0 & 57.3 & 74.2 & 77.9 & 66.9\\
            SHOT~\cite{SHOT}-Ens & \cmark & \xmark &58.6	&\textbf{25.2}	&55.3	&15.3	&\textbf{70.5}	&52.4	&46.2 & & 72.2 &59.3 &82.8 &82.9 &74.3\\
            DECISION~\cite{DECISION} & \cmark & \xmark &61.5	 &21.6	 &54.6 	 & {18.9}	 &67.5	&51.0	 &45.9 & & {74.5} &{59.4} &{84.4} & {83.6} &{75.5}\\
            CAiDA~\cite{caida} &\cmark & \xmark & - & - & - & - & - & - & - & & \textbf{75.2} & 60.5 & {84.7} & 84.2 & 76.2\\

            NRC~\cite{NRC}* & \cmark & \cmark & 65.8 & 24.1 & 56.0 & 16.0 & 69.2 & 53.4 & 47.4 & & 70.6 & 60.0 & 84.6 & 83.5 & 74.7 \\
            \rowcolor{gray!10} \textit{Ours (edge-mixup) + NRC} & \cmark & \cmark & 74.8 & 24.1 & 56.8 & 19.6 & 66.9 & 55.5 & 49.6 \textcolor{lightgreen}{(+2.2)} & & 72.1 & 62.9 & \textbf{86.4} & \textbf{84.8} & 76.6 \textcolor{lightgreen}{(+1.9)} \\
            
            \rowcolor{gray!10} \textit{Ours (feature-mixup) + NRC} & \cmark & \cmark &  \textbf{75.4} & 24.6 & \textbf{57.8} & \textbf{23.6} & 65.8 & \textbf{58.5} & \textbf{51.0} \textcolor{darkgreen}{\textbf{(+3.6)}} && 72.6 & \textbf{67.4} & 85.9 & 83.6 & \textbf{77.4} \textcolor{darkgreen}{\textbf{(+2.7)}} \\
            
            \bottomrule
            \end{tabular} 
        }
\vspace{-4mm}
\end{table*}

\begin{table*}[t]
    \centering
    \setlength{\tabcolsep}{3.5pt}
    \caption{\textbf{Single-Source DA (SSDA)} on Office-Home. SF indicates \textit{source-free}, \textcolor{darkgreen}{\textbf{(+x.x)}} indicates gains over NRC and SHOT++ respectively.}
    \label{tab:ssda_office_home}
    \resizebox{1\linewidth}{!}{%
        \begin{tabular}{lccccccccccccccl}
            \toprule
            \multirow{2}{30pt}{\centering Method} & \multirow{2}{*}{\centering SF} & \multicolumn{13}{c}{\textbf{Office-Home}
            } \\
            \cmidrule{4-16}
            && & {Ar}$\shortrightarrow${Cl} & {Ar}$\shortrightarrow${Pr} & {Ar}$\shortrightarrow${Rw} & {Cl}$\shortrightarrow${Ar} & {Cl}$\shortrightarrow${Pr} & {Cl}$\shortrightarrow${Rw} & {Pr}$\shortrightarrow${Ar} & {Pr}$\shortrightarrow${Cl} & {Pr}$\shortrightarrow${Rw} & {Rw}$\shortrightarrow${Ar} & {Rw}$\shortrightarrow${Cl} & {Rw}$\shortrightarrow${Pr} & Avg \\
            \midrule
			RSDA-MSTN~\cite{RSDA-MSTN} &\xmark& &53.2 &77.7 &81.3 &66.4 &74.0 &76.5 &67.9 &53.0 &82.0 &75.8 &57.8 &85.4 &70.9\\
			SENTRY~\cite{SENTRY} & \xmark && 61.8 & 77.4 & 80.1 & 66.3 & 71.6 & 74.7 & 66.8 & 63.0 & 80.9 & 74.0 & 66.3 & 84.1 & 72.2 \\ 
			FixBi~\cite{FixBi} &\xmark& & 58.1 & 77.3 & 80.4 & 67.7 & 79.5 & 78.1 & 65.8 & 57.9 & 81.7 & 76.4 & 62.9 & 86.7 & 72.7 \\
            SCDA~\cite{SCDA} & \xmark && 60.7 & 76.4 & 82.8 & 69.8 & 77.5 & 78.4 & 68.9 & 59.0 & 82.7 & 74.9 & 61.8 & 84.5 & 73.1 \\ 
			\midrule
			
	        SHOT~\cite{SHOT} &\cmark& &{57.1} & {78.1} & 81.5 & {68.0} & {78.2} & {78.1} & {67.4} & 54.9 & {82.2} & 73.3 & 58.8 & {84.3} & {71.8}  \\
		    
		    ${\text{A}^{2}\text{Net}}$~\cite{A2Net} &\cmark& &  58.4 & 79.0 & 82.4 & 67.5 & 79.3 & 78.9 & \textbf{68.0} & 56.2 & 82.9 & \textbf{74.1} & 60.5 & 85.0 & 72.8 \\
		    GSFDA~\cite{GSFDA} & \cmark && 57.9 & 78.6 & 81.0 & 66.7 & 77.2 & 77.2 & 65.6 & 56.0 & 82.2 & 72.0 & 57.8 & 83.4 & 71.3 \\ 
		    CPGA~\cite{CPGA} & \cmark && 59.3 & 78.1 & 79.8 & 65.4 & 75.5 & 76.4 & 65.7 & 58.0 & 81.0 & 72.0 & \textbf{64.4} & 83.3 & 71.6 \\ 
			{NRC}~\cite{NRC} &\cmark& &{57.7} 	&{80.3} 	&{82.0} 	&{68.1} 	&{79.8} 	&{78.6} 	&{65.3} 	&{56.4} 	&{83.0} 	&71.0	&{58.6} 	&{85.6} 	&{72.2} \\
            \rowcolor{gray!10} \textit{Ours (edge-mixup)} + NRC   &\cmark& &  60.8 & 80.1 & 81.6 & 67.2 & 79.3 & 78.5 & 65.4 & 61.0 & 83.8 & 70.2 & 63.1 & 85.3 & 73.0 \textcolor{lightgreen}{(+0.8)}   \\  
            
            \rowcolor{gray!10} \textit{Ours (feat-mixup)} + NRC   &\cmark& &  61.6 & 80.9 & 82.5 & 68.1 & 80.1 & 79.1 & 66.0 & \textbf{61.8} & 84.5 & 71.2 & 63.7 & 86.1 & 73.8  \textcolor{darkgreen}{\textbf{(+1.6)}}   \\       

		    SHOT++~\cite{SHOT++} & \cmark && 57.9 & 79.7 & 82.5 & 68.5 & 79.6 & 79.3 & 68.5 & 57.0 & 83.0 & 73.7 & 60.7 & 84.9 & 73.0 \\ 
            \rowcolor{gray!10} \textit{Ours (edge-mixup)} + SHOT++    &\cmark& &  61.0 & 80.4 & 82.1 & 67.6 & 79.8 & 78.8 & 67.1 & 60.7 & 84.3 & 73.0 & 63.5 & 85.7 & 73.7  \textcolor{lightgreen}{(+0.7)} \\
            
            \rowcolor{gray!10} \textit{Ours (feat-mixup)} + SHOT++   & \cmark & & \textbf{61.8} & \textbf{81.2} & \textbf{83.0} & \textbf{68.5} & \textbf{80.6} & \textbf{79.4} & 67.8 & 61.5 & \textbf{85.1} & 73.7 & 64.1 & \textbf{86.5} & \textbf{74.5}  \textcolor{darkgreen}{\textbf{(+1.5)}} \\
                        
            \bottomrule
            \end{tabular} 
        }
    \vspace{-4mm}
\end{table*}

\begin{table*}[t]
    \vspace{-2mm}
    \centering
    \caption{\textbf{Multi-Target Domain Adaptation (MTDA)} on Office-31. * indicates taken from~\citet{CGCT}}
    \label{supp:tab:mtda_office-31}
    \setlength{\tabcolsep}{9pt}
    \resizebox{0.9\linewidth}{!}{%
        \begin{tabular} {lcccccc}
        \toprule
        \multirow{2}{30pt}{\centering Method} & \multirow{2}{*}{\centering SF} & \multirow{2}{*}{\parbox{3cm}{\centering \vspace{4pt} w/o Domain \\ Labels}} & \multicolumn{4}{c}{\textbf{Office-31}}\\
        \cmidrule{4-7}
        & && Amazon$\shortrightarrow$ & DSLR$\shortrightarrow$ & Webcam$\shortrightarrow$ & Avg. \\
        \midrule
        Source train & \xmark & \cmark & 68.6 & 70.0 & 66.5 & 68.4 \\
        \midrule
        MT-MTDA~\cite{nguyen2020unsupervised} & \xmark & \xmark & 87.9 & 83.7 & 84.0 & 85.2 \\
        HGAN~\cite{yang2020heterogeneous} & \xmark & \xmark & 88.0 & 84.4 & 84.9 & 85.8 \\
        D-CGCT~\cite{CGCT} & \xmark & \xmark & {93.4} & {86.0} & {87.1} & {88.8} \\
        \midrule 
        JAN~\cite{long2017deep}* & \xmark & \cmark & 84.2 & 74.4 & 72.0 & 76.9 \\
        CDAN~\cite{long2018conditional}* & \xmark & \cmark & 93.6	& 80.5	& 81.3	& 85.1 \\
        AMEAN~\cite{chen2019blending} & \xmark & \cmark & 90.1 & 77.0 & 73.4 & 80.2 \\
        GDA~\cite{GDA}  & \xmark & \cmark & 88.8 & 74.5 & 73.2 & 87.9\\
        CGCT~\cite{CGCT}  & \xmark & \cmark & \textbf{93.9} & 85.1 & 85.6 & 88.2\\
        \rowcolor{gray!10}\textit{Ours (edge-mixup)}  & \cmark & \cmark & 90.3 & 87.1 & \textbf{86.9} & 88.1\\
        \rowcolor{gray!10}\textit{Ours (feature-mixup)}  & \cmark & \cmark & 92.5 & \textbf{88.4} & 86.5 & \textbf{89.1}\\
        \bottomrule
        \end{tabular}
        }
        \vspace{-2mm}
\end{table*}

\vspace{-1mm}
\noindent
\textbf{Datasets.}
We use four object classification DA benchmarks. {Office-31}~\cite{office-31} has three domains, 31 classes each: Amazon (\textbf{A}), DSLR (\textbf{D}), and Webcam (\textbf{W}).
{Office-Home}~\cite{office-home} contains four domains, 65 classes each: Artistic (\textbf{Ar}), Clipart (\textbf{Cl}), Product (\textbf{Pr}), and Real-world (\textbf{Rw}).
{VisDA}~\cite{visda} is a large-scale dataset with synthetic source and real target domains. {DomainNet}~\cite{M3SDA}
is the most challenging with six domains, 345 classes each: Clipart (\textbf{C}), Real (\textbf{R}), Infograph (\textbf{I}), Painting (\textbf{P}), Sketch (\textbf{S}), and Quickdraw (\textbf{Q}). For semantic segmentation DA, we use synthetic GTA5 (\textbf{G}) \cite{richter2016playing}, SYNTHIA (\textbf{Y}) \cite{ros2016synthia}, Synscapes (\textbf{S}) (\icmlcolor{Wrenninge et al.}, \citeyear{wrenninge2018synscapes}) as source datasets and real-world Cityscapes \cite{cordts2016cityscapes} as the target data. See {Appendix \ref{app:subsec:implementation}} for more details.

\vspace{-1mm}
\noindent
\textbf{Implementation details.} 
For object classification DA, we primarily use the source-free NRC \cite{NRC} for $\texttt{VsAlgo(.)}$ and $\texttt{CsAlgo(.)}$. Unless otherwise specified, \textit{Ours} implies NRC as $\texttt{VsAlgo(.)}$ and $\texttt{CsAlgo(.)}$. We follow NRC, with a ResNet-50 \cite{he2016deep_resnet} backbone for Office-Home, Office-31, and DomainNet, and a ResNet-101 for VisDA. For semantic segmentation DA, we follow \citet{kundu2021generalize} and use the standard DeepLabv2 \cite{chen2017deeplab} with a ResNet-101 backbone. We empirically set $\lambda\!=\!0.1$ for both edge and feature-mixup during both vendor-side and client-side training. We find that $\lambda\!=\!0.1$ works well across all settings and tasks. See {Appendix \ref{app:subsec:experimental_settings}} for extensive implementation details.

\begin{table}[t]
    \vspace{-4mm}
    \centering
    \caption{\textbf{Multi-Source DA (MSDA)} on Office-31. 
    }
    \vspace{0.5mm}
    \label{supp:tab:msda_office-31}
    \setlength{\tabcolsep}{5pt}
    \resizebox{\columnwidth}{!}{%
        \begin{tabular}{lccccc}
            \toprule
            \multirow{2}{*}{Method} & \multirow{2}{*}{SF} & \multicolumn{4}{c}{\textbf{Office-31}} \\
            \cmidrule{3-6}
            & & $\shortrightarrow$A & $\shortrightarrow$W & $\shortrightarrow$D & Avg. \\
            \midrule
            PFSA~\cite{PFSA} &\xmark  & 57.0 & 97.4 & 99.7 & 84.7 \\
            DCTN~\cite{DCTN} &\xmark  & 64.2  & 98.2 & 99.3  & 87.2 \\
            SImpAl~\cite{SImpAl} &\xmark  & 70.6 & 97.4 & 99.2 & 89.0 \\
            WAMDA~\cite{WAMDA} &\xmark  & 72.0 & 98.6 & 99.6 & 90.0 \\
            MFSAN~\cite{MFSAN} &\xmark  & {72.7} & {98.5}  & {99.5} & {90.2} \\
            MIAN~\cite{MIAN} & \xmark & 76.2 & 98.4 & 99.2 & 91.3 \\ 
            MLAN~\cite{MLAN} & \xmark & 75.7 & 98.8 & 99.6 & 91.4 \\ 
            \midrule
            {Source-combine} &\xmark  & 65.2 & 94.6 & 98.4 & 86.1\\
            SHOT\cite{SHOT}-Ens  &\cmark &{75.0} &{94.9} &{97.8} & {89.3}\\
            DECISION~\cite{DECISION} &\cmark & {75.4} &{98.4} &{99.6} & {91.1}\\
            CAiDA~\cite{caida} &\cmark & 75.8 & 98.9 & \textbf{99.8} & 91.6\\
            \rowcolor{gray!10}\textit{Ours (edge-mixup)} &\cmark & 76.3 & 99.1 & 99.4 & 91.6 \\
            \rowcolor{gray!10}\textit{Ours (feature-mixup)} &\cmark & \textbf{76.9} & \textbf{99.1} & 99.3 & \textbf{91.8}  \\
            \bottomrule
            \end{tabular} 
    }
    \vspace{-6mm}
\end{table}

\textbf{4.1. Comparison with prior arts}

\noindent
\textbf{a) Single Source Domain Adaptation (SSDA).}
Table \ref{tab:ssda_office_home} reports the results for object classification SSDA on Office-Home. Adding our proposed techniques to NRC \cite{NRC} improves NRC by 0.8\% for edge-mixup and 1.6\% for feature-mixup. Similarly, SHOT++ \cite{SHOT++} with our edge-mixup improves by {0.7\%} while feature-mixup gives an improvement of {1.5\%}. Further, we outperform even the non-source-free works on Office-Home. Table \ref{tab:ssda_office31_visda} shows the results on Office-31 and the VisDA dataset. Similar to Office-Home, we observe consistent improvements over both NRC and SHOT++ after adding feature-mixup ({1.1\%} and {1.5\%}) on Office-31. We also outperform the non-source-free works on the large-scale VisDA dataset.

\noindent
\textbf{b) Multi Source Domain Adaptation (MSDA).}
Table \ref{tab:msda_compare} shows the results for object classification MSDA on Office-Home and the large-scale DomainNet benchmark. We use our proposed mixup techniques with NRC, without using domain labels. We observe improvements of {1.9\%} for edge-mixup and {2.7\%} for feature-mixup on Office-Home. With the same settings, we observe consistent gains of {2.2\%} for edge-mixup and {3.6\%} for feature-mixup on DomainNet. In Table \ref{supp:tab:msda_office-31}, we compare our method to source-free and non-source-free prior works to assess our performance on Office-31 dataset. Even compared to non-source-free works, our technique obtains \textit{state-of-the-art} performance on the Office-31 benchmark despite not using domain labels.

\textbf{c) Multi-Target Domain Adaptation (MTDA).}
We adhere to the experimental protocols employed in \citet{chen2019blending,GDA}. We evaluate all possible combinations of one source domain and three target domains for MTDA for Office-31 benchmark in Table \ref{supp:tab:mtda_office-31}. Even without domain labels, our technique outperforms all others.

\textbf{d) DA for Semantic Segmentation.}
We use GtA \cite{kundu2021generalize} with our proposed mixup techniques for both SSDA and MSDA (Table~\ref{tab:daseg}). For edge-mixup, we obtain consistent gains of {1\%} and {1.2\%} over GtA on SSDA for GTA5$\to$Cityscapes (\textbf{G}) and SYNTHIA$\to$Cityscapes (\textbf{Y}) respectively. For MSDA with edge-mixup, we obtain consistent gains ({average 2.2\%}) over GtA on 4 settings (combinations of different source datasets). We also outperform the non-source-free SOTA MSDA-CL \cite{he2021multi} by an average {0.4\%}. In contrast to object classification DA, the gains of feature-mixup over GtA ({average 1.1\%}) are lower than those for edge-mixup. This is because feature-mixup better suits non-dense vectorized prediction tasks like object classification (where mixup is performed after the global average pooling layer). Whereas feature-mixup in semantic segmentation DA is performed on spatial conv.\ features.

\begin{figure*}[t]
\begin{center}
\centerline{\includegraphics[width=\textwidth]{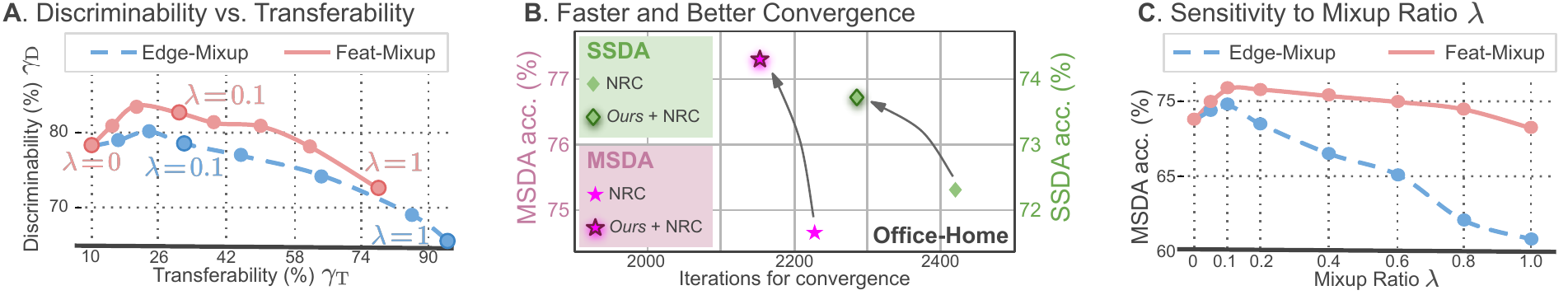}}
\vspace{-3mm}
\caption{\textbf{A.} Empirical discriminability vs. transferability for {Cl$\to$Rw (Office-Home)} at different mixup ratios $\lambda$. The plots are similar to the conceptual plot of Fig.\ \ref{fig:theoretical_insight2}\icmlcolor{B} and $\lambda\!=\!0.1$ presents a good trade-off. \textbf{B.} Faster and better convergence w.r.t. existing source-free works on both SSDA and MSDA for Office-Home, as mixup reduces the domain gap. \textbf{C.} Sensitivity to mixup ratio $\lambda$ for MSDA on Office-Home.}
\label{fig:analysis}
\end{center}
\vspace{-8mm}
\end{figure*}

\textbf{e) Faster and better convergence.} 
Fig.\ \ref{fig:analysis}\icmlcolor{B} shows the improved and faster convergence for our approach on both SSDA and MSDA on Office-Home. Since we effectively improve the transferability \ie reduce the domain gap with the mixup domains, the adaptation becomes faster. Further, as per Insight \icmlcolor{2}, the lowered target error upper bound leads to improved performance \ie better convergence.

\begin{table}[t]
\vspace{-4mm}
    \centering
    \caption{\textbf{Domain adaptation for semantic segmentation.} G, S, Y indicate GTA5, Synscapes, SYNTHIA datasets as source, SF indicates source-free, mIoU is 19-class for G column and 13-class for all others, \textcolor{darkgreen}{\textbf{(+x.x)}} indicates gains over prior source-free SOTA.}
    \vspace{1mm}
    \label{tab:daseg}
    \setlength{\tabcolsep}{2pt}
    \resizebox{1\columnwidth}{!}{%
        \begin{tabular}{lcccccccc}
            \toprule
            \multirow{2}{*}{\makecell[l]{Method \\ \textbf{($\boldsymbol{\to}$ Cityscapes)}}} & \multirow{2}{*}{SF} & \multicolumn{2}{c}{\textbf{SSDA}} & \multicolumn{5}{c}{\textbf{MSDA}} \\
            \cmidrule(l{3pt}r{3pt}){3-4} \cmidrule(l{3pt}r{3pt}){5-9}
            & & G & Y & G+S & S+Y & G+Y & G+S+Y & Avg \\
            \midrule
            FDA {\small (\icmlcolor{Yang et al.} \citeyear{yang2020fda})} & \xmark & 50.5 & 52.5 & -& -& -& - & - \\
            ProDA  {\small \cite{zhang2021prototypical}} & \xmark & 57.5 & 62.0 & - & - & - & - & - \\
            \midrule
            Src-combine {\small \cite{he2021multi}} & \xmark & - & - & 57.2 & 53.6 & 55.5 & 58.0 & 56.1 \\ 
            MSDA-CL {\small \cite{he2021multi}}    & \xmark & - & - &  \textbf{65.8} & 63.1 & 59.4 & \textbf{67.1} & 63.8 \\
            \midrule
            SFDA {\small \cite{liu2021source}} & \cmark & 43.1 & 45.9 & - & - & - & - & - \\
            URMA {\small (\icmlcolor{Teja et al.,} \citeyear{sivaprasad2021uncertainty})} & \cmark & 45.1 & 45.0 & - & - & - & - & - \\
            SFUDA {\small \cite{ye2021source}} & \cmark & 49.4 & 51.9 & - & - & - & - & - \\
            GtA {\small \cite{kundu2021generalize}}  & \cmark & 51.6 & 55.5 & 63.5 &  62.8 &  58.3 & 63.4 & 62.0 \\
            \rowcolor{gray!10} \textit{Ours (feature-mixup)} & \cmark & 51.9 &  55.6 & 63.6 & 63.2 & 61.4 & 64.3 & 63.1 \\
            \rowcolor{gray!10} &  & \textcolor{lightgreen}{{+0.3}} & \textcolor{lightgreen}{{+0.1}} & \textcolor{lightgreen}{{+0.1}} & \textcolor{lightgreen}{{+0.4}} & \textcolor{lightgreen}{{+3.1}} & \textcolor{lightgreen}{{+0.9}} & \textcolor{lightgreen}{{+1.1}} \\
            \rowcolor{gray!10} \textit{Ours (edge-mixup)} & \cmark & \textbf{52.6} & \textbf{56.7} & 64.6 & \textbf{65.4} & \textbf{61.8} & 64.9 & \textbf{64.2} \\
            \rowcolor{gray!10} &  & \textcolor{darkgreen}{\textbf{+1.0}} & \textcolor{darkgreen}{\textbf{+1.2}} & \textcolor{darkgreen}{\textbf{+1.1}} & \textcolor{darkgreen}{\textbf{+2.6}} & \textcolor{darkgreen}{\textbf{+3.5}} & \textcolor{darkgreen}{\textbf{+1.5}} & \textcolor{darkgreen}{\textbf{+2.2}} \\
            \bottomrule        
        \end{tabular} 
        }
\vspace{-4mm}
\end{table}

\textbf{4.2. Analysis}

\textbf{a) Effect of sub-domain augmentations.}
We performed an ablation study for SSDA on Office-Home (see Table~\ref{tab:supp_ablation_office_home}) to disentangle the gains from the use of sub-domain augmentations in feature-mixup. While sub-domain augmentations improve NRC and SHOT++ by {0.6\%} and {0.5\%} respectively, we observe that feature-mixup provides further improvements of {$\sim$1\%} in both cases. Thus, the gains from feature-mixup can be attributed more to the proposed mixup than the sub-domain augmentations.

\textbf{b) Empirical transferability vs. discriminability.}
Fig.\ \ref{fig:analysis}\icmlcolor{A} illustrates the empirical curve between discriminability and transferability for different $\lambda$ in edge-mixup and feature-mixup. Transferability $\gamma_\mathrm{T}$ is evaluated on mixup domain data with Eq.\ \ref{eqn:target_risk_bound}, \ref{eqn:transf_disc_defn} using a domain-classifier trained on original source-target data. Discriminability $\gamma_\mathrm{D}$ is evaluated with Eq.\ \ref{eqn:target_risk_bound}, \ref{eqn:transf_disc_defn} by computing the accuracy of a source-target joint supervised task classifier. Both domain classifier and task classifier are trained on the features of a frozen backbone $h$ trained with $\texttt{VsAlgo}(\mathcal{D}_{s_m})$ with a particular $\lambda$. The empirical plots are similar to the conceptual plot of Fig.\ \ref{fig:theoretical_insight2}\icmlcolor{B}.

\begin{table}[t]
\vspace{-4mm}
    \centering
    \setlength{\tabcolsep}{12pt}
    \caption{
    \textbf{Compatibility with non-source-free DA} works on Office-Home. SSDA and MSDA indicate single-source and multi-source DA. \textcolor{darkgreen}{\textbf{(+x.x)}} indicates gains over corresponding baseline.
    }
    \label{tab:nonsf_comparison}
    \resizebox{\columnwidth}{!}{
    \begin{tabular}{lll}
        \toprule
        \multirow{2}{*}{Method} &  \multicolumn{2}{c}{\textbf{Office-Home}} \\
        \cmidrule(lr){2-3}
         & SSDA & MSDA \\
        \midrule
        DANN \cite{ganin2016domain}  & 57.6 & 64.6\\
        DANN + feature-mixup  & \textbf{67.2} \textcolor{darkgreen}{\textbf{(+9.6)}} & \textbf{72.9} \textcolor{darkgreen}{\textbf{(+8.3)}} \\
        \midrule
        CDAN+E \cite{long2018conditional} &  65.8 & 69.4\\
        CDAN+E + feature-mixup & \textbf{72.2} \textcolor{darkgreen}{\textbf{(+6.4)}} & \textbf{74.1} \textcolor{darkgreen}{\textbf{(+4.7)}}\\
        \midrule
        SRDC \cite{SRDC}  & {71.3} & 73.1\\
        SRDC + feature-mixup &  \textbf{72.3} \textcolor{darkgreen}{\textbf{(+1.0)}} & \textbf{75.1} \textcolor{darkgreen}{\textbf{(+2.0)}}\\
        \bottomrule
    \end{tabular}%
    }
\vspace{-4mm}
\end{table}

\textbf{c) Sensitivity to mixup ratio $\boldsymbol{\lambda}$.}
We conduct a sensitivity analysis for mixup ratio $\lambda$ (Fig.\ \ref{fig:analysis}\icmlcolor{C}) for MSDA on Office-Home. For edge-mixup, we observe gains over the baseline ($\lambda\!=\!0$) until $\lambda\!=\!0.1$, followed by a drop \wrt the baseline for $\lambda \!>\! 0.2$. For higher $\lambda$, the proportion of edges increases. The significant loss of texture and color causes a sizeable drop in discriminability (Fig.\ \ref{fig:analysis}\icmlcolor{A}), leading to poor adaptation performance. In contrast, the flexibility of feature-space mixup enables it to outperform the baseline consistently upto $\lambda\!=\!0.8$. Thus, a relatively low $\lambda$ is preferable for edge-mixup while feature-mixup is more invariant to $\lambda$.

\textbf{d) Qualitative analysis for semantic segmentation DA.} Figure \ref{fig:qualitative_analysis} illustrates qualitative results on Cityscapes (target dataset) for different SSDA (G) and MSDA settings (G+S, S+Y, G+Y, G+S+Y) for semantic segmentation. We compare with the vendor-side source-only baseline and the prior source-free SOTA \cite{kundu2021generalize} and highlight the improvement regions with white circles.

\textbf{4.3. Compatibility with non-source-free DA}

Table \ref{tab:nonsf_comparison} examines the complementary nature of the proposed mixup to prior non-source-free SSDA works on Office-Home. For MSDA comparisons, multiple sources are merged into a single source. For SSDA, adding feature-mixup enhances DANN by 9.8\%, CDAN by 6.4\% and SRDC by 1\%. Similarly, for MSDA, adding feature-mixup improves DANN by {8.3\%}, CDAN by {4.7\%} and SRDC by 2\%. These consistent improvements demonstrate our general compatibility with non-source-free DA works.

\begin{table}[t]
\vspace{-4mm}
    \centering
    \caption{
         \textbf{Ablation study on sub-domain augmentations} for SSDA on Office-Home benchmark. \textcolor{darkgreen}{\textbf{(+x.x)}} indicates improvements over NRC and SHOT++ respectively.
    }
    \label{tab:supp_ablation_office_home}
    \setlength{\tabcolsep}{25pt}
    \resizebox{1\columnwidth}{!}{
        \begin{tabular}{ll}
            \toprule
            Method &  Average Acc. \\
            \midrule
			{NRC}~\cite{NRC} & {72.2} \\
			NRC + sub-domain augs.  & 72.8  \textcolor{lightgreen}{{(+0.6)}} \\
            \rowcolor{gray!10} \textit{Ours (feat-mixup)} + NRC   &  \textbf{73.8}  \textcolor{darkgreen}{\textbf{(+1.6)}}   \\
            \midrule
		    SHOT++~\cite{SHOT++}  &  73.0 \\ 
            SHOT++ + sub-domain augs. &  73.5  \textcolor{lightgreen}{{(+0.5)}} \\
            \rowcolor{gray!10} \textit{Ours (feat-mixup)} + SHOT++    & \textbf{74.5}  \textcolor{darkgreen}{\textbf{(+1.5)}} \\
            \bottomrule
        \end{tabular} 
    }  
\end{table}

\section{Conclusion}
We study the perspective of discriminability-transferability trade-off in source-free DA. Identifying the key hurdles to extending available trade-off-based techniques, we investigate the idea of generic domain representations without concurrent source-target access. Based on our insights, we propose novel ways to realize generic domains, but observe degraded task-discriminability. As a remedy, we operate on intermediate mixup domains and theoretically demonstrate that the mixup domains achieve a tighter bound on target error, thereby improving the DA performance. This simple modification, added to prior DA approaches, yields state-of-the-art performance across source-free benchmarks for single-source and multi-source DA on both classification and segmentation. Since the procedure for realizing a generic-domain is somewhat task-dependent, future work can focus on learnable realizations of generic-domains.

\textbf{Acknowledgements.}
This work was supported by MeitY (Ministry of Electronics and Information Technology) project (No. 4(16)2019-ITEA), Govt. of India.

{
\small
\bibliography{ref}
\bibliographystyle{icml2022}
}

\newpage
\appendix
\onecolumn

{\Large \textbf{Appendix}}

\begin{figure*}[ht]
\begin{center}
\centerline{\includegraphics[width=\textwidth]{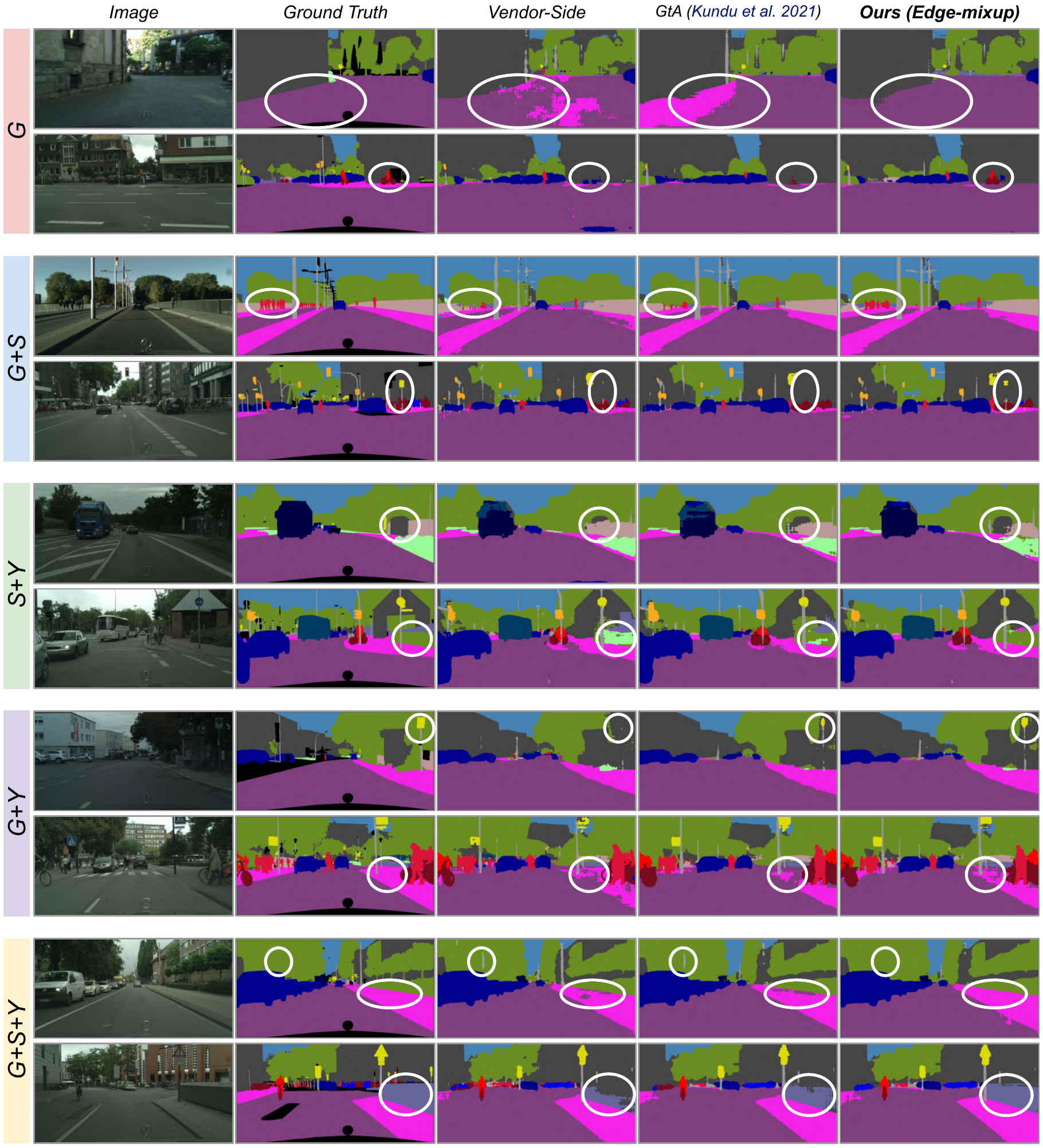}}
\vspace{-3mm}
\caption{Qualitative analysis on target Cityscapes validation set for semantic segmentation SSDA (G) and MSDA (G+S, S+Y, G+Y, G+S+Y) settings. White circles indicate areas of improvement w.r.t. prior art and vendor-side baseline. \textit{Best viewed in color}.}
\label{fig:qualitative_analysis}
\end{center}
\vspace{-8mm}
\end{figure*}

In this appendix, we provide more details of our approach and theoretical insights, extensive implementation details, additional quantitative and qualitative performance analysis and ablation studies. Towards reproducible research, we will publicly release our complete codebase and trained network weights.
The appendix is organized as follows:

\vspace{-2mm}

\begin{itemize}
\setlength{\itemindent}{-3mm}
\setlength{\itemsep}{-1mm}
    \item Section~\ref{app:sec:notations}: Notations (Table~\ref{sup:tab:notations})
    \item Section~\ref{app:sec:theory}: Discussions related to theory
    \item Section~\ref{app:sec:experiments}: Experiments
    \vspace{-2mm}
    \begin{itemize}
        \setlength{\itemindent}{-3mm}
        \item Implementation details (Sec. \ref{app:subsec:implementation})
        \item Experimental settings (Sec.~\ref{app:subsec:experimental_settings})
        \item Additional results (Sec.~\ref{app:subsec:add_results}, Table \ref{tab:speech_text}, Fig.\ \ref{fig:qualitative_analysis})
        \item Extended comparisons (Sec.~\ref{app:subsec:extended_comp}, Table \ref{supp:tab:ssda_office_home}, \ref{supp:tab:ssda_office31_visda}, \ref{supp:tab:msda_compare})
    \end{itemize}
\end{itemize}{}

\section{Notations}
\label{app:sec:notations}

We summarize the notations used throughout the paper in Table \ref{sup:tab:notations}. The notations are listed under 7 groups \ie models, distributions, theory-related, datasets, samples, spaces and miscellaneous.

\begin{table}[h]
    \centering
    \caption{\textbf{Notation Table}}
    \label{sup:tab:notations}
    \vspace{1mm}
    \setlength{\tabcolsep}{25pt}
    \resizebox{0.7\columnwidth}{!}{%
        \begin{tabular}{lcl}
        \toprule
        
         & \multirow{1}{*}{Symbol} & \multirow{1}{*}{Description} \\
         \midrule

         \multirow{3}{*}{\rotatebox[origin=c]{0}{Models}} & $h$ & Backbone feature extractor  \\
         & $f_c$ & Task classifier \\
         & $f_d$ & Domain classifier \\
         \midrule

        \multirow{6}{*}{\rotatebox[origin=c]{0}{Distributions}} & $p_s$ & Source marginal distribution \\
         & $p_t$ & Target marginal distribution \\
         & $p_{s_g}$ & Source-generic marginal distr. \\
        & $p_{t_g}$ & Target-generic marginal distr. \\
         & $p_{s_m}$ & Source-mixup marginal distr. \\
        & $p_{t_m}$ & Target-mixup marginal distr. \\
        \midrule

        \multirow{4}{*}{\rotatebox[origin=c]{0}{Theory-related}} & $\epsilon_s$ & Expected source risk \\
         & $\epsilon_t$ & Expected target risk \\
         & $d_{\mathcal{H}}$ & $\mathcal{H}$-divergence \\
         & $\mathcal{H}$ & Backbone hypothesis space \\

        \midrule
        \multirow{4}{*}{\rotatebox[origin=c]{0}{Datasets}} & $\mathcal{D}_s$ & Labeled source dataset  \\
        & $\mathcal{D}_{s_m}$ & Labeled source-mixup dataset  \\
        & $\mathcal{D}_t$ & Unlabeled target dataset  \\
        & $\mathcal{D}_{t_m}$ & Unlabeled target-mixup dataset  \\
        \midrule
        \multirow{9}{*}{\rotatebox[origin=c]{0}{Samples}} & $(x_s, y_{s})$ & Labeled source sample  \\
        & $x_{s_g}$ & Generic domain sample of $x_s$  \\
        & $(x_{s_m}, y_{s})$ & Labeled source-mixup sample  \\
        & $x_t$ & Unlabeled target sample  \\
        & $x_{t_m}$ & Unlabeled target-mixup sample \\
        & $z_s$ & Features of sample $x_s$ \\
        & $z_s^{[i]}$ & Features of sample with $i^\text{th}$ aug. \\
        & $z_{s_g}$ & Generic domain features of $z_s$ \\
        & $z_{s_m}$ & Source-mixup features \\
         \midrule 
        \multirow{3}{*}{\rotatebox[origin=c]{0}{Spaces}} 
        & $\mathcal{X}$ & Input space \\
        & $\mathcal{Z}$ & Backbone feature space \\
        & $\mathcal{C}$ & Label set for goal task \\
        \midrule
        \multirow{2}{*}{\rotatebox[origin=c]{0}{Miscellaneous}}  & $\mathcal{A}_e$ & Edge estimation method \\
        & $\mathcal{A}^{[i]}$ & $i^\text{th}$ augmentation function \\
        \bottomrule
        \end{tabular}
        }
\end{table}

\section{Discussions related to theory}
\label{app:sec:theory}

\subsection{Theorem \icmlcolor{1} and Proof}
\label{app:subsec:theorem1_proof}

\textbf{Theorem 1.\ (Mixup $\boldsymbol{\mathcal{H}}$-divergence)}
\textit{Assume that original source $p_s$ and target $p_t$ are easily separable} \ie \textit{perfect accuracy for domain classifier $f_d$. Also assume that source-generic $p_{s_g}$ and target-generic $p_{t_g}$ domains are impossible to separate} \ie \textit{accuracy of $f_d$ imitates that of a random classifier. For a linear domain classifier $f_d$;}

\vspace{-4mm}
\begin{equation}
    d_\mathcal{H}(p_{s_m}, p_{t_m})\leq d_\mathcal{H}(p_{s}, p_{t})
    \label{eqn:app_hdiv_bound}
\end{equation}
\vspace{-6mm}

\textit{Proof.}
From Eq. \ref{eqn:target_risk_bound},

\vspace{-8mm}
\begin{align}
\begin{split}
    d_\mathcal{H}(p_s,p_t)\!&=\!\sup_{h' \in \mathcal{H}} \left|\expectation_{x\sim p_s}[\mathbbm{1}(f_d(z) \!=\! 1)] \!-\! \expectation_{x\sim p_t}[\mathbbm{1}(f_d (z)\! =\! 1)]\right| \\
    &= \sup_{h' \in \mathcal{H}} \left| \phi(p_s) - \phi(p_t) \right|; \text{ where } z = h'(x)
\end{split}
\label{eqn:app_dh_defn}
\end{align}
\vspace{-4mm}

Note that domain classifier produces 0 for source and 1 for target. Thus, $\phi(p_s)$ is the source-domain-classification \textit{error} while $\phi(p_t)$ is the target-domain-classification \textit{accuracy} since the conditions in both indicator functions are for 1 (\ie target). Recall that $p_{s_g}$ and $p_{t_g}$ are the source and target generic distributions respectively.

First, we focus on $\phi(p_{s_m})$ \ie source-domain-classification error for the source-mixup distribution $p_{s_m}$. We replace the condition of $f_d(z)=1$ to $f_d(z)>0$ since $f_d$ would practically be a binary domain classifier relying on sigmoid activation, but we require $f_d$ to simply be a linear function without any non-linear activation function. Then,

\vspace{-5mm}
\begin{align*}
\phi(p_{s_m}) = \expectation_{x \sim p_s} \left[ \mathbbm{1}\left(f_d\left(\lambda z_{s_g} + (1-\lambda) z\right) >0 \right) \right]; \text{ where } z = h(x)
\end{align*}

We assume that $f_d$ is a linear function \ie $f_d(ka+b)=k f_d(a) + f_d(b)$ for a scalar constant $k$. And we use the property that expectation over indicator function is a probability,

\vspace{-5mm}
\begin{align*}
\phi(p_{s_m}) = \Pr_{x \sim p_s} \left[ \lambda f_d(z_{s_g})+(1-\lambda)f_d(z) >0 \right]
\end{align*}

Now we consider the possible separate cases where $\lambda f_d(z_{s_g})+(1-\lambda)f_d(z) >0$ is satisfied (as shown in the Table below).

\begin{table}[!h]
    \centering
    \setlength{\tabcolsep}{10pt}
    \resizebox{0.6\columnwidth}{!}{    
    \begin{tabular}{lccl}
    \toprule
         & $\lambda f_d(z_{s_g})$ & $(1-\lambda)f_d(z)$ & Comments \\ \midrule
        1 & $>0$ & $>0$ & $>0$ always \\ \midrule
        2.1 & $>0$ & $<0$ & $>0$ if $\lambda \vert f_d(z_{s_g})\vert>(1-\lambda)\vert f_d(z)\vert$ \\
        2.2 & $>0$ & $<0$ & $<0$ otherwise \\ \midrule
        3.1 & $<0$ & $>0$ & $>0$ if $\lambda \vert f_d(z_{s_g})\vert<(1-\lambda)\vert f_d(z)\vert$ \\
        3.2 & $<0$ & $>0$ & $<0$ otherwise \\ \midrule
        4 & $<0$ & $<0$ & $<0$ always \\ \bottomrule
    \end{tabular}
    }
\end{table}

Using 1, 2.1, 3.1 from the table above, we split the probability term into a sum of product of probabilities of the three cases,

\vspace{-5mm}
\begin{gather*}
    \phi(p_{s_m}) = \Pr_{x \sim p_s}[\lambda f_d(z_{s_g}) > 0]\Pr_{x \sim p_s}[(1-\lambda)f_d(z)>0] + \\ \Pr_{x \sim p_s}[\lambda f_d(z_{s_g}) > 0]\Pr_{x \sim p_s}[(1-\lambda)f_d(z)<0]\Pr_{x \sim p_s}\left[\lambda \vert f_d(z_{s_g})\vert>(1-\lambda)\vert f_d(z)\vert \right] + \\
\Pr_{x \sim p_s}[\lambda f_d(z_{s_g}) < 0]\Pr_{x \sim p_s}[(1-\lambda)f_d(z)>0]\Pr_{x \sim p_s}\left[\lambda\vert f_d(z_{s_g})\vert<(1-\lambda)\vert f_d(z)\vert \right]
\end{gather*}

Note that $\lambda$ and $(1-\lambda)$ terms do not matter when the product of $\lambda$ or $(1-\lambda)$ with $f_d(.)$ is compared with $0$, as both are positive scalar constants. Further, using $\phi(p_s)\!=\!\Pr_{x \sim p_s} [f_d(z) \!>\! 0]$ and $\phi(p_{s_g}) \!=\! \Pr_{x \sim p_s}[f_d(z_{s_g}) \!>\! 0]$, we get,

\vspace{-5mm}
\begin{gather*}
    \phi(p_{s_m}) = \phi(p_{s_g})\phi(p_s) + \phi(p_{s_g})(1-\phi(p_s))\Pr_{x \sim p_s}\left[\lambda f_d(z_{s_g})>(1-\lambda)f_d(z) \right] + \\
(1-\phi(p_{s_g}))\phi(p_s)\Pr_{x \sim p_s}\left[\lambda f_d(z_{s_g})<(1-\lambda)f_d(z) \right]
\end{gather*}

To simplify, we define $\zeta_s = \Pr_{x \sim p_s}\left[\lambda f_d(z_{s_g})>(1-\lambda)f_d(z) \right]$ where $\zeta_s \in [0, 1]$. Note that the other probability term is $(1-\zeta_s)$ due to the complement rule of probabilities.

\vspace{-5mm}
\begin{gather*}
    \phi(p_{s_m}) = \phi(p_{s_g})\phi(p_s) + \phi(p_{s_g})(1-\phi(p_s))\zeta_s + (1-\phi(p_{s_g}))\phi(p_s)(1-\zeta_s)
\end{gather*}

Now, we use the assumption that the source-generic domain $p_{s_g}$ and target-generic domain $p_{t_g}$ are impossible to separate \ie accuracy of $f_d$ imitates that of a random classifier. Formally, $\phi(p_{s_g})$ represents the source-domain-classification error for the source-generic domain. Thus, $\phi(p_{s_g})\to \frac{1}{2}$ (since binary classification). Using this and simplifying,

\vspace{-5mm}
\begin{align}
    \begin{split}
        \phi(p_{s_m}) = \phi(p_s)(1-\zeta_s)+\frac{\zeta_s}{2}
    \end{split}
\end{align}

Similarly, we can derive the same expression for the target-mixup domain \ie $\phi(p_{t_m})$ except that the assumption of target-domain-classification \textit{accuracy}, imitating the accuracy of a random classifier, for target-generic domain has to be used \ie $\phi(p_{t_g})\to \frac{1}{2}$. We get,

\vspace{-5mm}
\begin{align}
    \begin{split}
        \phi(p_{t_m}) = \phi(p_t)(1-\zeta_t)+\frac{\zeta_t}{2}   
    \end{split}
\end{align}

We know that $0\leq \zeta_s, \zeta_t \leq 1$ since both are probabilities. Further, we had assumed that original source and target domains are easily separable \ie perfect accuracy (or zero error) for domain classifier. Thus, $\phi(p_s)\to 0$ and $\phi(p_t) \to 1$ as per their interpretations of error and accuracy respectively. With these constraints on $\phi(p_s), \phi(p_t), \zeta_s, \zeta_t$, it can be easily shown that,

\vspace{-5mm}
\begin{align}
    \begin{split}
        \phi(p_{s_m})\geq \phi(p_s)\; \forall\; \zeta_s \in [0, 1];\;\;\;\; \phi(p_{t_m}) \leq \phi(p_t) \;\forall\; \zeta_t \in [0, 1]
    \end{split}
    \label{eqn:app_proof_prefinal}
\end{align}

The interpretation is that source-domain-classification error increases while target-domain-classification accuracy decreases after mixup. In other words, the mixup domain features are more domain-invariant compared to the original domain features. Note that while $\lambda$ does not directly appear in Eq. \ref{eqn:app_proof_prefinal}, both $\zeta_s$ and $\zeta_t$ vary with $\lambda$. Thus, $\lambda$ will control both the decrease in accuracy and the increase in error \ie the $\mathcal{H}$-divergence relation between mixup domains and original domains is not independent of $\lambda$.

Now, we add the two inequalities of Eq. \ref{eqn:app_proof_prefinal} and rearrange the terms,

\vspace{-5mm}
\begin{gather*}
    \phi(p_s) + \phi(p_{t_m}) \leq \phi(p_{s_m}) + \phi(p_t) \\
\phi(p_{t_m}) - \phi(p_{s_m}) \leq \phi(p_t) - \phi(p_s)
\end{gather*}

Finally, applying absolute value function on both sides (without sign change as both sides are positive) and taking supremum on both sides over $h \in \mathcal{H}$,

\vspace{-5mm}
\begin{gather*}
    \sup_{h\in \mathcal{H}}\vert\phi(p_{t_m}) - \phi(p_{s_m})\vert \leq \sup_{h\in \mathcal{H}}\vert\phi(p_t) - \phi(p_s)\vert
\end{gather*}

Applying the definition of $\mathcal{H}$-divergence from Eq. \ref{eqn:app_dh_defn}, we arrive at our result \ie Eq. \ref{eqn:app_hdiv_bound}. \hfill $\square$

\subsection{Regarding result relating $\kappa(p_{s_m}, p_{t_m})$ and $\kappa(p_{s}, p_{t})$}
\label{app:subsec:mixupk_relation}

As noted in the main paper, the assumptions on $f_c$ (involved in the computation of $\kappa(.,.)$ in Eq.\ \ref{eqn:target_risk_bound}) become unrealistic if we use a similar strategy as in the proof of Theorem \icmlcolor{1}. Concretely, we may assume $f_c$ to be a linear classifier in order to separate the mixup terms over $f_c$. However, we cannot make realistic assumptions on the source-generic error $\epsilon_{s_g}(h)$ or target-generic error $\epsilon_{t_g}(h)$ in the way that assumptions were made for the domain classifier $f_d$. This is because the realizable generic domains have high transferability by definition, but lack a strict condition on their discriminability (\ie on $\epsilon_{s_g}(h)$ or $\epsilon_{t_g}(h)$).
However, we observe in practice that the discriminability is preserved over a reasonable range of $\lambda$ (see Fig.\ \ref{fig:analysis}\icmlcolor{A}).

\section{Experiments}
\label{app:sec:experiments}
\subsection{Implementation details}
\label{app:subsec:implementation}
\noindent
\textbf{Datasets.}
We assess the efficacy of our approach on four object classification DA benchmarks. The \textbf{Office-31}~\cite{office-31} benchmark contains three domains from office settings: Amazon (\textbf{A}), DSLR (\textbf{D}), and Webcam (\textbf{W}), each with 31 object categories. \textbf{Office-Home}~\cite{office-home}, a more complex dataset, consists of images of everyday objects from four domains, each having 65 classes: Artistic (\textbf{Ar}), Clipart (\textbf{Cl}), Product (\textbf{Pr}), and Real-world (\textbf{Rw}). \textbf{VisDA}~\cite{visda} is a large-scale dataset with 152,397 synthetic images in the source domain and 55,388 real-world images in the target domain. Finally, \textbf{DomainNet}~\cite{M3SDA} is the most challenging due to its extremely diversified domains, high class imbalance, and 6 domains with 345 classes each: Clipart (\textbf{C}), Real (\textbf{R}), Infograph (\textbf{I}), Painting (\textbf{P}), Sketch (\textbf{S}), and Quickdraw (\textbf{Q}).

\subsection{Experimental settings}
\label{app:subsec:experimental_settings}

For object classification DA, we primarily use the source-free NRC \cite{NRC} for $\texttt{VsAlgo(.)}$ and $\texttt{CsAlgo(.)}$. We follow NRC, with a ResNet-50 \cite{he2016deep_resnet} backbone for Office-Home, Office-31, and DomainNet, and a ResNet-101 backbone for VisDA. For edge-mixup, we choose a CNN-based edge detector \cite{soria2020dexined}, pretrained on a dataset with images irrelevant to the usual DA benchmark datasets. We use the Adam optimizer \cite{kingma2014adam} with a learning rate of 1e-3, momentum of 0.9, and batch size of 64 for training with label smoothing following \cite{NRC,SHOT}. For semantic segmentation DA, we follow \citet{kundu2021generalize} and use the standard DeepLabv2 \cite{chen2017deeplab} with a ResNet-101 backbone. We omit the cPAE as it presents an extra model that complicates the training process and is orthogonal to our contributions.

\noindent
\paragraph{Sub-domain augmentations.}
We selected to depict the sub-domains using the following four strong augmentations (total 5 sub-domains) on the original data:

{a) FDA}: In this augmentation \cite{yang2020fda}, the image is stylized based on a reference image by exchanging the low frequency FFT spectrum. We use images from a style transfer dataset (\icmlcolor{Huang et al.,} \citeyear{huang2017adain}) as reference images.

{b) Weather augmentation}: We employ \citet{imgaug}'s frost (weather condition) augmentation. There are five severity levels, and we randomly choose one from the lowest three to govern the augmentation strength.

{c) AdaIN}: By altering the convolutional feature statistics in an instance normalization layer (\icmlcolor{Ulyanov et al.}, \citeyear{ulyanov2017improved}), this approach (\icmlcolor{Huang et al.,} \citeyear{huang2017adain}) stylizes the image depending on a reference image. To guarantee adequate content retention, we set the stylization strength to $0.3$ (on a scale of 0 to 1, \ie original image to severely stylized). We use images from a style transfer dataset (\icmlcolor{Huang et al.,} \citeyear{huang2017adain}) as reference images.

{d) Cartoonization}: We employ the \citet{imgaug} cartoonization augmentation. The strength of the augmentation has no controllable parameter.

\noindent
\paragraph{Feature-mixup training.}
We use the same network architecture as NRC \cite{NRC}, substituting the classifier with a fully connected layer with batch normalization (\icmlcolor{Ioffe et al.}, \citeyear{ioffe2015batch}) and a fully connected layer with weight normalization (\icmlcolor{Salimans et al.}, \citeyear{salimans2016weight}). We divide the network into a backbone $h$ (upto ResLayer-4 followed by Global Average Pooling) and the classifier $f_c$, as defined in Sec.\ \ref{sec:approach}. First, the network is initialized from a pre-trained model on the source dataset $\mathcal{D}_s$ with random sub-domain augmentations. During vendor-side and client-side training, the mixup operation is performed between the random augmentation feature $z$ extracted by the backbone and the corresponding domain generic feature $z_{g}$ obtained by averaging over the sub-domain augmentation features as demonstrated in Fig.\ \ref{fig:approach}\icmlcolor{B}. The obtained mixup feature $z_m$ is then passed through the classifier $f_c$ and the loss computation is done as per the predefined \texttt{VsAlgo(.)} and \texttt{CsAlgo(.)}. 

\noindent
\paragraph{Edge-mixup training.}
In contrast, the edge-mixup optimization algorithm is much simpler. Here, the RGB image $x$ is directly mixed-up with the edge representation of the image \ie $\mathcal{A}_e(x)$ extracted using an edge estimation model~\cite{soria2020dexined}, with the mixup ratio $\lambda$. The mixup image with enhanced edge features is directly used as input for the vendor-side and client-side training.

\subsection{Additional results}
\label{app:subsec:add_results}

\begin{wraptable}[7]{r}{0.62\textwidth}
    \vspace{-7.5mm}
    \caption{
         {Evaluation on Libri-Adapt benchmark \cite{mathur2020libri} and Amazon reviews benchmark \cite{blitzer2007biographies}.}
    }
    \label{tab:speech_text}
    \setlength{\tabcolsep}{5pt}
    \resizebox{\linewidth}{!}{
    \begin{tabular}{cc}
        \begin{tabular}{lc}
        \toprule
          
        \multicolumn{2}{c}{\begin{tabular}{c}\textbf{DA for Speech Recog. (audio)} \\on Libri-Adapt dataset \end{tabular}} \\ \midrule
        CMatch \cite{Hou2021CrossdomainSR} & 71.8 \\
        \rowcolor{gray!10} + \textit{feat-mixup} & \textbf{73.5} \\
        \bottomrule
        \end{tabular} 
        &
        \begin{tabular}{lc}
        \toprule
        \multicolumn{2}{c}{\begin{tabular}{c}\textbf{DA for Sentiment Clsf. (text)} \\ on Amazon Reviews dataset\end{tabular}}\\ \midrule
        UDALM \cite{Karouzos2021UDALMUD} & 91.7\\
        \rowcolor{gray!10} + \textit{feat-mixup} & \textbf{93.1} \\
        \bottomrule
        \end{tabular} 
    \end{tabular}
    }
\end{wraptable}

\textbf{Evaluation on other modalities like text and speech.}
Our key theoretical insights build on data-type-agnostic DA theory \cite{ben2006analysis} and are useful for any ML data. While \textit{contours} are vision-specific, similar generic-domains can be formed using \textit{task-specific} knowledge (e.g. via FFT for speech). Moreover, our feature-mixup relies on \textbf{a)} deeper layer feature space and \textbf{b)} augmentations which are prevalent in DL for any data-type and task. We demonstrate the gains of feature-mixup on audio DA and text DA benchmarks (Table \ref{tab:speech_text}) using open source audio\footnote{\url{https://github.com/iver56/audiomentations}} and speech\footnote{\url{https://github.com/makcedward/nlpaug}} augmentations.

\subsection{Extended comparisons}
\label{app:subsec:extended_comp}

We show extended comparisons with more prior arts due to lack of space in the main paper, but the main results are the same. Table \ref{supp:tab:ssda_office_home} shows SSDA for Office-Home (Table \ref{tab:ssda_office_home} in the main paper). Table \ref{supp:tab:ssda_office31_visda} shows SSDA for Office-31 and VisDA (Table \ref{tab:ssda_office31_visda} in the main paper). Finally, Table \ref{supp:tab:msda_compare} shows MSDA on DomainNet and Office-Home (Table \ref{tab:msda_compare} in the main paper).

\begin{table*}[h]
    \centering
    \setlength{\tabcolsep}{4pt}
    \caption{Single-Source Domain Adaptation (SSDA) on Office-Home benchmarks. SF indicates \textit{source-free} adaptation.}
    \label{supp:tab:ssda_office_home}
    \resizebox{1\linewidth}{!}{%
        \begin{tabular}{lccccccccccccccc}
            \toprule
            \multirow{2}{30pt}{\centering Method} & \multirow{2}{*}{\centering SF} & \multicolumn{13}{c}{\textbf{Office-Home}} \\
            \cmidrule{4-16}
            && & {Ar}$\shortrightarrow${Cl} & {Ar}$\shortrightarrow${Pr} & {Ar}$\shortrightarrow${Rw} & {Cl}$\shortrightarrow${Ar} & {Cl}$\shortrightarrow${Pr} & {Cl}$\shortrightarrow${Rw} & {Pr}$\shortrightarrow${Ar} & {Pr}$\shortrightarrow${Cl} & {Pr}$\shortrightarrow${Rw} & {Rw}$\shortrightarrow${Ar} & {Rw}$\shortrightarrow${Cl} & {Rw}$\shortrightarrow${Pr} & Avg \\
            \midrule
			SRDC \cite{SRDC} &\xmark& &52.3 & 76.3 & {81.0} & {69.5} & {76.2} & {78.0} & {68.7} & {53.8} & {81.7} & {76.3} & {57.1} & {85.0} & {71.3} \\
			SENTRY~\cite{SENTRY} & \xmark && 61.8 & 77.4 & 80.1 & 66.3 & 71.6 & 74.7 & 66.8 & 63.0 & 80.9 & 74.0 & 66.3 & 84.1 & 72.2 \\ 
            TCM~\cite{TCM} & \xmark && 58.6 & 74.4 & 79.6 & 64.5 & 74.0 & 75.1 & 64.6 & 56.2 & 80.9 & 74.6 & 60.7 & 84.7 & 70.7 \\ 
            CDS~\cite{CDS} & \xmark && 55.9 & 74.6 & 77.7 & 65.5 & 73.3 & 75.0 & 67.8 & 54.5 & 79.5 & 73.7 & 59.3 & 82.4 & 69.9 \\ 
            SCDA~\cite{SCDA} & \xmark && 60.7 & 76.4 & 82.8 & 69.8 & 77.5 & 78.4 & 68.9 & 59.0 & 82.7 & 74.9 & 61.8 & 84.5 & 73.1 \\ 
			\midrule
	        SHOT~\cite{SHOT} &\cmark& &{57.1} & {78.1} & 81.5 & {68.0} & {78.2} & {78.1} & {67.4} & 54.9 & {82.2} & 73.3 & 58.8 & {84.3} & {71.8}  \\
		    ${\text{A}^{2}\text{Net}}$~\cite{A2Net} &\cmark& &  58.4 & 79.0 & 82.4 & 67.5 & 79.3 & 78.9 & \textbf{68.0} & 56.2 & 82.9 & \textbf{74.1} & 60.5 & 85.0 & 72.8 \\
		    GSFDA~\cite{GSFDA} & \cmark && 57.9 & 78.6 & 81.0 & 66.7 & 77.2 & 77.2 & 65.6 & 56.0 & 82.2 & 72.0 & 57.8 & 83.4 & 71.3 \\ 
		    CPGA~\cite{CPGA} & \cmark && 59.3 & 78.1 & 79.8 & 65.4 & 75.5 & 76.4 & 65.7 & 58.0 & 81.0 & 72.0 & \textbf{64.4} & 83.3 & 71.6 \\ 
		    SFDA~\cite{SFDA}  & \cmark && 48.4 & 73.4 & 76.9 & 64.3 & 69.8 & 71.7 & 62.7 & 45.3 & 76.6 & 69.8 & 50.5 & 79.0 & 65.7 \\ 
			{NRC}~\cite{NRC} &\cmark& &{57.7} 	&{80.3} 	&{82.0} 	&{68.1} 	&{79.8} 	&{78.6} 	&{65.3} 	&{56.4} 	&{83.0} 	&71.0	&{58.6} 	&{85.6} 	&{72.2} \\
            \rowcolor{gray!10} \textit{Ours (edge-mixup)} + NRC   &\cmark& &  60.8 & 80.1 & 81.6 & 67.2 & 79.3 & 78.5 & 65.4 & 61.0 & 83.8 & 70.2 & 63.1 & 85.3 & 73.0   \\  
            \rowcolor{gray!10} \textit{Ours (feat-mixup)} + NRC   &\cmark& &  61.6 & 80.9 & 82.5 & 68.1 & 80.1 & 79.1 & 66.0 & \textbf{61.8} & 84.5 & 71.2 & 63.7 & 86.1 & 73.8   \\       
		    SHOT++~\cite{SHOT++} & \cmark && 57.9 & 79.7 & 82.5 & 68.5 & 79.6 & 79.3 & 68.5 & 57.0 & 83.0 & 73.7 & 60.7 & 84.9 & 73.0 \\ 
            \rowcolor{gray!10} \textit{Ours (edge-mixup)} + SHOT++    &\cmark& &  61.0 & 80.4 & 82.1 & 67.6 & 79.8 & 78.8 & 67.1 & 60.7 & 84.3 & 73.0 & 63.5 & 85.7 & 73.7 \\
            \rowcolor{gray!10} \textit{Ours (feat-mixup)} + SHOT++   & \cmark & & \textbf{61.8} & \textbf{81.2} & \textbf{83.0} & \textbf{68.5} & \textbf{80.6} & \textbf{79.4} & 67.8 & 61.5 & \textbf{85.1} & 73.7 & 64.1 & \textbf{86.5} & \textbf{74.5} \\
            \bottomrule
            \end{tabular} 
        }
\end{table*}

\begin{table*}[h]
    \centering
    \setlength{\tabcolsep}{12pt}
    \caption{\textbf{Single-Source Domain Adaptation (SSDA)} on Office-31 and VisDA benchmarks. SF indicates \textit{source-free} adaptation.}
    \label{supp:tab:ssda_office31_visda}
    \resizebox{\textwidth}{!}{
    \begin{tabular}{lccccccccccc}
        \toprule
        \multirow{2}{20pt}{\centering Method}& \multirow{2}{*}{\centering SF} 
        &&  \multicolumn{7}{c}{\textbf{Office-31}} && \multicolumn{1}{c}{\textbf{VisDA}} \\
        \cmidrule(lr){4-10}\cmidrule(lr){12-12}
        &&& A$\rightarrow$D & A$\rightarrow$W & D$\rightarrow$W & W$\rightarrow$D & D$\rightarrow$A & W$\rightarrow$A & Avg && S $\rightarrow$ R \\
        \midrule
		FixBi~\cite{FixBi} & \xmark& & 95.0 & 96.1 & 99.3 & 100.0 & 78.7 & 79.4 & 91.4 && 87.2 \\
		FAA~\cite{FAA} & \xmark && 94.4 & 92.3 & 99.2 & 99.7 & 80.5 & 78.7 & 90.8 && 82.7\\ 
		CDAN+RADA~\cite{RADA} & \xmark && 96.1 & 96.2 & 99.3 & 100.0 & 77.5 & 77.4 & 91.1 && 76.3\\ 
		RFA~\cite{RFA} & \xmark && 93.0 & 92.8 & 99.1 & 100.0 & 78.0 & 77.7 & 90.2 && 79.4\\ 
		SCDA~\cite{SCDA} & \xmark && 95.4 & 95.3 & 99.0 & 100.0 & 77.2 & 75.9 & 90.5 && -\\ 
		\midrule
		SHOT~\cite{SHOT}& \cmark& & {94.0} &  90.1 &  98.4 & {99.9}  & 74.7 & 74.3 & {88.6} && 82.9\\
		3C-GAN~\cite{3C-GAN}& \cmark& & {92.7} &  {93.7} &  98.5 & 99.8  & {75.3} & {77.8} & {89.6} && 81.6\\
		CPGA~\cite{CPGA} & \cmark && 94.4 & 94.1 & 98.4 & 99.8 & 76.0 & 76.6 & 89.9 && 84.1\\ 
		HCL~\cite{HCL} & \cmark && 90.8 & 91.3 & 98.2 & 100.0 & 72.7 & 72.7 & 87.6 && 83.5\\ 
		SFDA~\cite{SFDA} & \cmark && 92.2 & 91.1 & 98.2 & 99.5 & 71.0 & 71.2 & 87.2 && 76.7\\ 
		VDM-DA~\cite{VDM-DA}  & \cmark && 93.2 & 94.1 & 98.0 & 100.0 & 75.8 & 77.1 & 89.7 && 85.1\\ 
		${\text{A}^{2}\text{Net}}$~\cite{A2Net} & \cmark && 94.5 & \textbf{94.0} & 99.2 & 100.0 & 76.7 & 76.1 & 90.1 && 84.3\\ 
		NRC~\cite{NRC} & \cmark && 96.0 & 90.8 & 99.0 & 100.0 & 75.3 & 75.0 & 89.4 && 85.9\\ 
		\rowcolor{gray!10} \textit{Ours (edge-mixup)} + NRC & \cmark && 96.1 & 92.4 & 99.2 & 100.0 & 76.9 & 77.1 & 90.3 && 86.4 \\
		\rowcolor{gray!10} \textit{Ours (feat-mixup)} + NRC & \cmark && \textbf{96.3} & 92.8 & \textbf{99.2} & 100.0 & 77.4 & 77.5 & 90.5 && {87.3} \\
		SHOT++~\cite{SHOT++} & \cmark && 94.3 & 90.4 & 98.7 & 99.9 & 76.2 & 75.8 & 89.2 && {87.3}\\ 
		\rowcolor{gray!10} \textit{Ours (edge-mixup)} + SHOT++ & \cmark && 94.4 & 92.0 & 98.9 & 100.0 & 77.8 & 77.9 & 90.2 && 	87.5 \\
		\rowcolor{gray!10} \textit{Ours (feat-mixup)} + SHOT++ & \cmark && 94.6 & 93.2 & 98.9 & \textbf{100.0} & \textbf{78.3} & \textbf{78.9} & \textbf{90.7} && 	\textbf{87.8} \\
        \bottomrule
    \end{tabular}
    }
    \vspace{-4mm}
\end{table*}



\begin{table*}[t]
    \centering
    \caption{\textbf{Multi-Source Domain Adaptation (MSDA)} on DomainNet and Office-Home. SF indicates \textit{source-free} adaptation. The middle section compares SSDA works by adapting each source-target pair and reporting the best and worst results for each target.
    }
    \label{supp:tab:msda_compare}
    \setlength{\tabcolsep}{6pt}
    \resizebox{1\linewidth}{!}{%
        \begin{tabular}{lcccccccccccccccc}
            \toprule
            \multirow{2}{30pt}{\centering Method} & \multirow{2}{*}{\centering SF} & \multirow{2}{*}{\parbox{3cm}{\centering \vspace{4pt} w/o Domain \\ Labels}} & \multicolumn{7}{c}{\textbf{DomainNet}} && \multicolumn{5}{c}{\textbf{Office-Home}} \\
            \cmidrule(lr){4-10} \cmidrule(lr){12-16}
            & & &$\shortrightarrow$C & $\shortrightarrow$I & $\shortrightarrow$P & $\shortrightarrow$Q & $\shortrightarrow$R & $\shortrightarrow$S & Avg & & $\shortrightarrow$Ar & $\shortrightarrow$Cl & $\shortrightarrow$Pr & $\shortrightarrow$Rw & Avg \\
            \midrule
            {MDAN \cite{MDAN}} & \xmark & \xmark & 52.4 & 21.3 & 46.9 & 8.6 & 54.9 & 46.5 & 38.4 && 68.1 & 67.0 & 81.0 & 82.8 & 74.7 \\ 
            {SImpAl$_{50}$~\cite{SImpAl}} & \xmark & \xmark &   66.4 & 26.5 & 56.6 & 18.9 & 68.0 & 55.5 & 48.6 & & 70.8 & 56.3 & 80.2 & 81.5 & 72.2 \\ 
            {CMSDA~\cite{CMSDA}} & \xmark & \xmark &  70.9 & 26.5 & 57.5 & 21.3 & 68.1 & 59.4 & 50.4 & &  71.5 & 67.7 & 84.1 & 82.9 & 76.6 \\ 
            DARN~\cite{DARN} &\xmark & \xmark  &- &- &- &- &- & - & - && 70.0 & 68.4 & 82.7 & 83.9 & 76.3 \\
            {DRT \cite{DRT}} & \xmark & \xmark &   71.0 & 31.6 & 61.0 & 12.3 & 71.4 & 60.7 & 51.3 &  & - & - & - & - & -\\ 
            STEM~\cite{STEM} & \xmark & \xmark & 72.0 & 28.2 & 61.5 & 25.7 & 72.6 & 60.2 & 53.4 & & - & - & - & - & -  \\ 
            \midrule
             SHOT\cite{SHOT}-worst & \cmark  & \xmark &14.8	&1.0	&3.5	&3.8	&6.6	&11.9	&7.0 & & 66.6	&53.8	&77.9	&80.8 &69.8\\ 
             SHOT\cite{SHOT}-best & \cmark  & \xmark&58.3	&22.7	&53.0	&18.7	&65.9	&48.4	&44.5 & & 72.1	&57.2	&83.4	&81.3 &73.5\\ 
            \rowcolor{gray!10} \textit{Ours}-worst  & \cmark & \xmark & 17.0 & 2.4 & 8.7 & 4.9 & 12.1 & 18.1 & 10.5 && 66.0 & 61.6 & 80.1 & 79.1 & 71.7\\
            \rowcolor{gray!10} \textit{Ours}-best  & \cmark& \xmark & 68.6 & 21.9 & 53.8 & 17.4 & 65.4 & 58.1 & 47.5 && 71.2 & 63.7 & 86.1 & 84.5 & 76.4\\
            \midrule
            {Source-combine} &\xmark & \cmark &57.0 &23.4 &54.1 &14.6 &67.2 &50.3 & 44.4 & & 58.0 & 57.3 & 74.2 & 77.9 & 66.9\\
            SHOT~\cite{SHOT}-Ens & \cmark & \xmark &58.6	&25.2	&55.3	&15.3	&\textbf{70.5}	&52.4	&46.2 & & 72.2 &59.3 &82.8 &82.9 &74.3\\
            DECISION~\cite{DECISION} & \cmark & \xmark &61.5	 &21.6	 &54.6 	 &\textbf{18.9}	 &67.5	&51.0	 &45.9 & & {74.5} &{59.4} &{84.4} & {83.6} &{75.5}\\
            NEL~\cite{NEL} & \cmark & \xmark & 68.3 & 22.1 & 54.7 & 22.8 & 67.3 & 57.1 & 48.7 & & - & - & - & - & - \\ 
            CAiDA~\cite{caida} &\cmark & \xmark & - & - & - & - & - & - & - & & 75.2 & 60.5 & 84.7 & 84.2 & 76.2\\
            \rowcolor{gray!10} \textit{Ours (edge-mixup) + NRC} & \cmark & \cmark & 74.8 & 24.1 & 56.8 & 19.6 & 66.9 & 55.5 & 49.6 & & 72.1 & 62.9 & \textbf{86.4} & \textbf{84.8} & 76.6 \\
            \rowcolor{gray!10} \textit{Ours (feature-mixup) + NRC} & \cmark & \cmark &  \textbf{75.4} & 24.6 & \textbf{57.8} & \textbf{23.6} & 65.8 & \textbf{58.5} & \textbf{51.0} && 72.6 & \textbf{67.4} & 85.9 & 83.6 & \textbf{77.4} \\
            \bottomrule
            \end{tabular} 
        }
    \vspace{-4mm}
\end{table*}

\end{document}